\documentclass[10pt,twocolumn,letterpaper]{article}

\usepackage{cvpr}              
\usepackage{bm}
\usepackage{booktabs}
\usepackage{multirow}
\usepackage{siunitx}
\usepackage{footmisc}
\usepackage[table]{xcolor}
\definecolor{bestcolor}{RGB}{255,200,200}  
\definecolor{secondcolor}{RGB}{255,255,200}
\definecolor{thirdcolor}{RGB}{200,255,200}

\definecolor{cvprblue}{rgb}{0.21,0.49,0.74}
\usepackage[pagebackref,breaklinks,colorlinks,allcolors=cvprblue]{hyperref}

\def\paperID{****} 
\def\confName{CVPR}
\def\confYear{2026}

\title{GaussianPile: A Unified Sparse Gaussian Splatting Framework for Slice-based Volumetric Reconstruction}

\author{Di Kong$^{1,2}$ ~~~~~ Yikai Wang$^{3}$ ~~~~~ Wenjie Guo$^{1}$ ~~~~~ Yifan Bu$^{1}$ ~~~~~ Boya Zhang$^{2,4}$ ~~~~~ Yuexin Duan$^{2,4}$ \\ Xiawei Yue$^{2,4}$ ~~~~~ Wenbiao Du$^{2,5}$ ~~~~~ Yiman Zhong$^{2,6}$ ~~~~~ Yuwen Chen$^{1}$\textsuperscript{\dag} ~~~~~ Cheng Ma$^{1,2}$\textsuperscript{\dag}\\
{$^{1}$Tsinghua University~~~~~$^{2}$Zhongguancun Academy~~~~~$^{3}$Beijing Normal University}\\
{$^{4}$Nankai University~~~~~$^{5}$Beijing Institute of Technology~~~~~$^{6}$Beihang University}\\
{\tt\small {kd24@mails.tsinghua.edu.cn}; {chen-yw97@mail.tsinghua.edu.cn}; {cheng\_ma@tsinghua.edu.cn}}}

\begin{document}
\maketitle
{
\renewcommand{\thefootnote}{\dag}
\footnotetext{Corresponding to Cheng Ma and Yuwen Chen.}
}
\begin{abstract}
Slice-based volumetric imaging is widely applied and it demands representations that compress aggressively while preserving internal structure for analysis. We introduce GaussianPile, unifying 3D Gaussian splatting with an imaging system-aware focus model to address this challenge. Our proposed method introduces three key innovations: (i) a slice‑aware piling strategy that positions anisotropic 3D Gaussians to model through‑slice contributions, (ii) a differentiable projection operator that encodes the finite‑thickness point spread function of the imaging acquisition system, and (iii) a compact encoding and joint optimization pipeline that simultaneously reconstructs and compresses the Gaussian sets. Our CUDA-based design retains the compression and real‑time rendering efficiency of Gaussian primitives while preserving high‑frequency internal volumetric detail. Experiments on microscopy and ultrasound datasets demonstrate that our method reduces storage and reconstruction cost, sustains diagnostic fidelity, and enables fast 2D visualization, along with 3D voxelization. In practice, it delivers high-quality results in as few as $3$ minutes, up to $11\times$ faster than NeRF-based approaches, and achieves consistent $16\times$ compression over voxel grids, offering a practical path to deployable compression and exploration of slice-based volumetric datasets.
\end{abstract}
\section{Introduction}
\label{sec:intro}

\begin{figure}[!t]
\centerline{\includegraphics[width=\columnwidth]{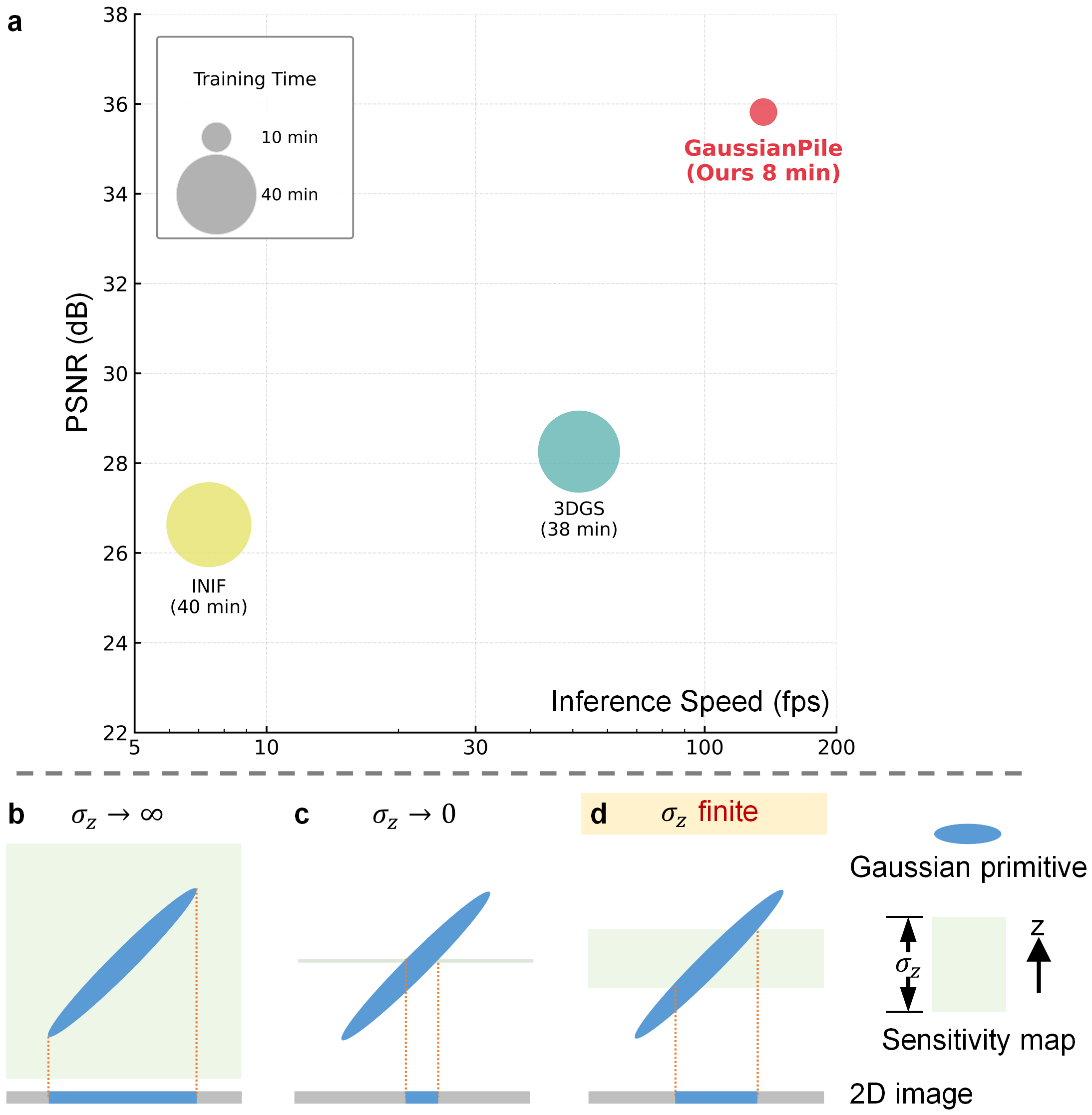}}
\vspace{-2pt}
\caption{Panel (a) is the average PSNR-fps-minute comparison (circle radius encodes minutes of training). Our method achieves the highest accuracy at far lower compute cost than prior work. Panels (b–d) are the comparisons of Gaussian rendering models under different imaging physics. (b) All-in-focus ($\sigma_z \to \infty$): no axial falloff; primitives contribute regardless of its depth, appropriate for all-in-focus pinhole rendering (e.g., original 3DGS \cite{kerbl20233d}) or line-integral modalities (e.g., X-ray \cite{cai2024radiative}). (c) Zero-thickness ($\sigma_z \to 0$): delta-like axial response; primitives contribute only at the exact plane, suitable for dense slicing (e.g., MRI \cite{liang2025innergs}). (d) Finite-thickness Focus Gaussian (ours): finite axial sensitivity ($\sigma_z$ finite); primitives contribute within a focal zone determined by $\sigma_z$, modeling anisotropic PSF in slice-based modalities (e.g., ultrasound, light-sheet microscopy).}
\label{FIG_define_formulation}
\vspace{-6pt}
\end{figure}

The rapid evolution of modern imaging technologies, particularly in biomedical and scientific domains, has led to an exponential growth in data volume \cite{peng2014virtual, long20093d}. 
High-resolution, multi-dimensional imaging, such as 3D microscopy and volumetric ultrasound, generates datasets that pose a formidable challenge for storage, transmission, and subsequent analysis. 
The substantial cost associated with managing this data deluge is, in turn, becoming a limiting factor that constrains the broader application of these advanced imaging techniques.

Classical image and video codecs, including the widely adopted JPEG \cite{wallace1991jpeg} and HEVC \cite{sze2014high} standards, offer partial relief. 
However, they are fundamentally designed for 2D imagery or temporal sequences of frames, lacking specific optimization for the inherent redundancy and structural characteristics of 3D volumetric data. 
Consequently, their application to such data often results in suboptimal compression ratios or the loss of scientifically critical information \cite{dong2015compression}.

The rise of artificial intelligence has opened new avenues for data compression \cite{yang2020improving, kingma2013auto}. 
A promising paradigm leverages Implicit Neural Representations (INR) \cite{sitzmann2020implicit}, where a compact neural network learns a continuous mapping from spatial coordinates to signal values (e.g., pixel intensity). 
The continuity of the activation functions allows the network to represent complex data with a remarkably small number of parameters, thereby achieving high compression ratios (CR). 
Despite this promise, INR-based approaches face significant hurdles: they are prone to losing high-frequency details, and their iterative training and inference processes are computationally intensive and time-consuming, especially for large-scale datasets \cite{xu2023nesvor, liang2022coordx}. 
This often necessitates a patch-based compression strategy and relegates them primarily to a "cold storage" scenario, where rapid access and real-time interaction are not feasible.

Recently, 3D Gaussian Splatting (3DGS) has emerged as a groundbreaking technique for novel view synthesis, demonstrating unparalleled efficiency in both reconstruction and rendering \cite{kerbl20233d}. 
By representing a scene as a collection of anisotropic 3D Gaussians, 3DGS achieves real-time rendering with stunning visual quality. 
The continuity of the Gaussian function ensures that the number of primitives required is far less than the number of original voxels, implying an inherent compression capability. 
However, the standard 3DGS framework is fundamentally designed to model surface appearance from multi-view images. 
It discards internal volumetric information, making it unsuitable for direct application to scientific and medical imaging, where the complete internal structure of a subject is of paramount importance.

To bridge this critical gap, we introduce GaussianPile, a novel volumetric representation paradigm that synergizes the strengths of both approaches (high CR of INR and fast fitting of 3DGS). We re-engineer the core rendering principle of Gaussian primitives by incorporating the physical properties of the imaging system. This method preserves the high compression ratio and computational efficiency of Gaussian-based representations while addressing the previously unresolved case of slice-based volumetric imaging with finite focal depth, a regime that dominates ultrasound, light-sheet microscopy (LSM), and structured illumination microscopy (SIM) applications. It enables not only efficient storage but also fast reconstruction and real-time rendering of the internal structures, facilitating interactive exploration and analysis that was previously impractical. As shown in Fig.~\ref{FIG_define_formulation} (a), this physics-grounded design yields the best PSNR–speed–training-time balance among all compared methods.

\section{Related Work}
\label{sec:relatedwork}

\subsection{INR and Biomedical Image Compression}
INR have emerged in recent years as a novel paradigm for compressing high-dimensional biomedical images \cite{mildenhall2021nerf, zha2022naf, cai2024structure, zhu2025implicit, di2025survey}. 
By storing lightweight network weights instead of the massive original voxel data, this family of methods theoretically enables extremely high compression ratios \cite{yang2024sharing, dai2025implicit}, and has been explored through neural sparse-grid representation \cite{lu2021compressive}, accelerated coordinate-based model \cite{han2022coordnet}, divide-and-conquer neural fields \cite{han2025dcinr}, as well as tensor-based factorization \cite{ballester2019tthresh}.
Although strategies like inter-patch weight sharing can partially accelerate training, their compression efficiency still lags orders of magnitude behind traditional codecs (e.g., HEVC \cite{sze2014high}) \cite{ma2024semantic}.
These factors collectively limit the widespread application of INR in scenarios requiring rapid compression or real-time interaction.
This motivates explicit, rasterization-friendly representations that preserve volumetric fidelity while supporting efficient rendering and compression.

\subsection{3DGS for Volumetric Data Modeling}
3DGS \cite{kerbl20233d} represents scenes or objects using trainable Gaussian primitives and achieves real-time rendering through parallel rasterization. 
It has been widely adopted for scene modeling \cite{wu20244d, yang2024deformable, yu2024mip, zhang2024gaussian}, 3D generation approaches incorporating diffusion models \cite{ren2023dreamgaussian4d, xu2024grm, chen2024text} and medical imaging \cite{cai2024radiative,zha2024r,yu2025x,liang2025innergs}, etc. Furthermore, by applying techniques such as quantization, clustering, spatial structure, and dynamic optimization, the storage requirements have been significantly reduced while maintaining the rendering quality \cite{lee2024compact, bagdasarian20253dgs, wang2025compressing, niedermayr2024compressed}.
To extend the concept of 3DGS into the domain of volumetric data, recent research has begun exploring modifications to the rendering module to adapt to different slice formation physics. 
For instance, in integral-projection-based modalities (e.g., CBCT,  Fig. \ref{FIG_define_formulation}  (b)), where 2D image intensity reflects line integrals of contributions along the infinite projection rays, adjustments to rendering equations have enabled successful adaptation \cite{cai2024radiative, gao2024ddgs}. 
Conversely, for dense-slice-based modalities (e.g., MRI,  Fig. \ref{FIG_define_formulation}  (c)), where the value of each 2D image is determined solely by the signal at its corresponding spatial location (theoretically zero slice thickness), exploration has also been conducted \cite{liang2025innergs}.

Nevertheless, for the prevalent intermediate case where the slice thickness is a finite value—corresponding to imaging systems with an anisotropic point spread function, such as light-sheet microscopy, structured illumination super-resolution imaging, and ultrasound—the rendering model for Gaussian primitives remains an open problem that has not been adequately resolved (Fig. \ref{FIG_define_formulation}  (d)). 
Existing methods struggle to accurately model this finite focal zone: naively applying 3DGS yields plausible 2D slice renderings but produces incoherent 3D structures with severe floating artifacts, as the primitives lack physical constraints on their axial extent and thus contribute indiscriminately across slices. This constitutes a key obstacle to the successful application of Gaussian splatting techniques for general volumetric data compression.

\section{Preliminary: 3D Gaussian Splatting}

\begin{figure}[!t]
\centerline{\includegraphics[width=\columnwidth]{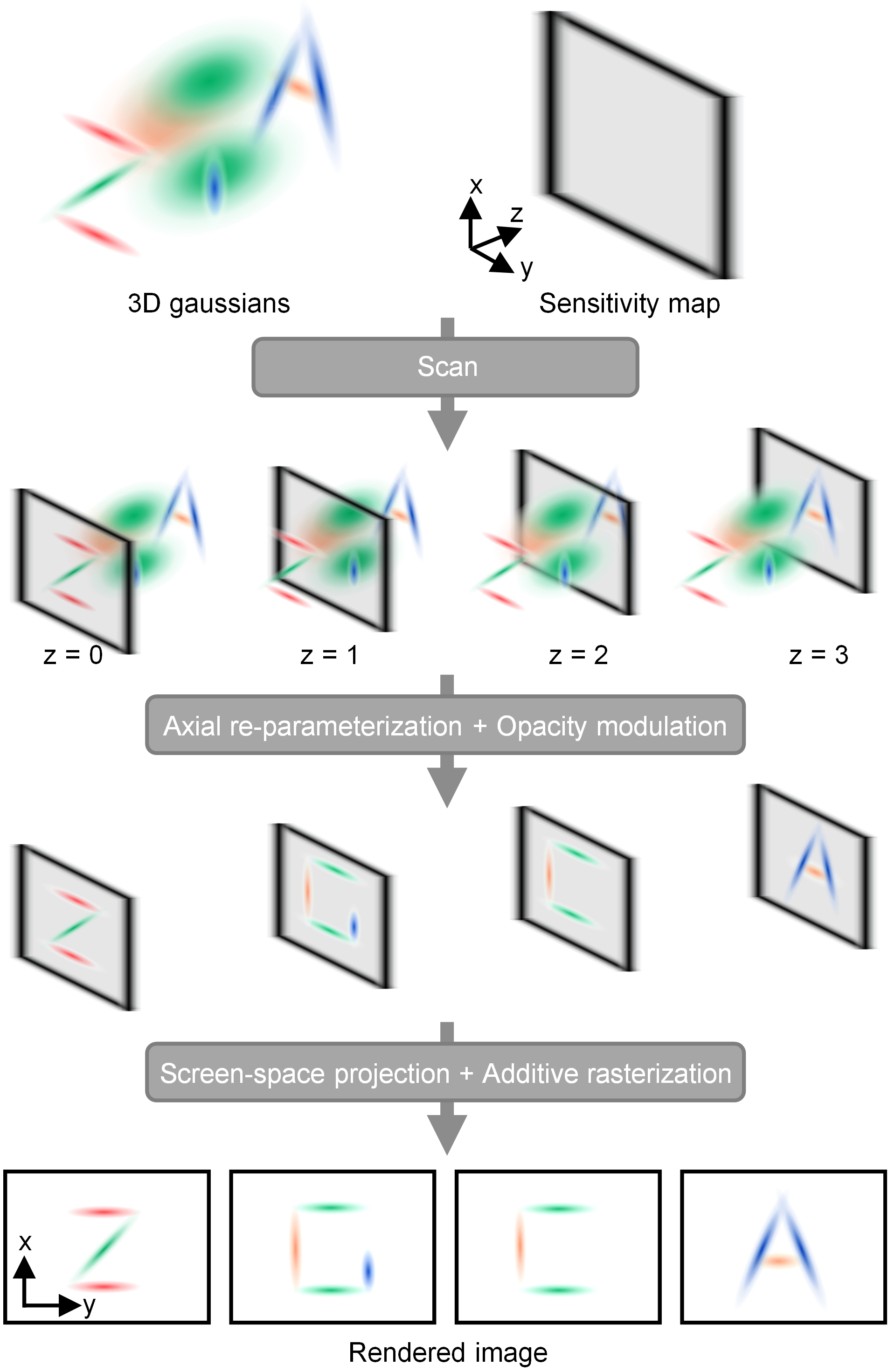}}
\caption{Focus-aware rendering pipeline of GaussianPile. Given 3D Gaussians and a sensitivity map defining the focal zone, the rendering process consists of three stages: \textbf{(1) Scan}: project Gaussians onto slices at different depths; \textbf{(2) Axial re-parameterization and Opacity modulation}: apply axial weighting to attenuate off-focal contributions, yielding Focus Gaussians, while modulating opacity based on distance from the focal plane; \textbf{(3) Screen-space projection and Additive rasterization}: compute 2D marginal distributions and additively accumulate weighted Gaussian footprints to form final rendered images.}
\label{FIG_render_diagram}
\end{figure}

In 3D Gaussian Splatting, a 3D scene is represented as a collection of anisotropic Gaussian primitives distributed in space.
Each Gaussian defines a continuous volumetric density with theoretically infinite support, whose contribution at a spatial location $\mathrm{\mathbf{x}} \in \mathbb{R}^3$ is described by an unnormalized Gaussian function:
\begin{equation}
p(\mathrm{\mathbf{x}}) = \exp\left(-\tfrac{1}{2}(\mathrm{\mathbf{x}}-\boldsymbol{\mu})^{\top}\boldsymbol{\Sigma}^{-1}(\mathrm{\mathbf{x}}-\boldsymbol{\mu})\right),
\end{equation}
where $\boldsymbol{\mu} = (\mu_x, \mu_y, \mu_z)^{\top}$ denotes the Gaussian center. The covariance matrix $\boldsymbol{\Sigma}$ encodes the spatial extent and orientation of the Gaussian and is factorized as $\boldsymbol{\Sigma} = \bm{R}\bm{S}\bm{S}^{\top}\bm{R}^{\top}$, where $\bm{S} = \mathrm{diag}(s_x, s_y, s_z)$ represents anisotropic scaling and $\bm{R}$ is a rotation matrix derived from a unit quaternion $\bm{q}$.

Each Gaussian further carries learnable spherical harmonics (SH) coefficients to model view-dependent color, and an opacity parameter $\alpha$ controlling its blending contribution along the rendering ray. All parameters — including position, scale, orientation, color, and opacity — are jointly optimized via gradient-based rendering loss minimization, enabling efficient and differentiable volumetric representation. Based on this representation, our method extends 3DGS to slice-based tomographic reconstruction.

\section{Method}

GaussianPile bridges 3D Gaussian Splatting and volumetric slice-based reconstruction through a physics-aware focus model, as shown in Fig.~\ref{FIG_render_diagram}. In this section, we first introduce our design of Focus-aware physical model in Sec.~\ref{modeling}. We then derive new forward rendering in Sec.~\ref{pile}, implemented with differentiable CUDA kernels, maintaining real-time efficiency.  The optimization strategy is elaborated in Sec.~\ref{optimize}.

\subsection{Focus-aware Modeling}
\label{modeling}

The fundamental innovation of GaussianPile lies in incorporating the physical model of focused imaging into the rendering process. To facilitate computational convenience, we conduct this analysis within the camera coordinate system, with relevant parameters distinguished by the subscript $c$. Additionally, we assume that the point spread function (PSF) of the imaging system, $\text{psf}(\mathbf{x}_c)$, is a spatially invariant 3D Gaussian function: 

\begin{equation}
\text{psf}(\mathbf{x}_c) \propto
\exp\left( -\frac{1}{2} \left( \frac{x_c^2}{\sigma_x^2} + \frac{y_c^2}{\sigma_y^2} + \frac{z_c^2}{\sigma_z^2} \right) \right),
\end{equation}
where $\sigma_x$,  $\sigma_y$ and  $\sigma_z$ correspond to the spatial resolution of the imaging system in the x, y, and z directions, respectively. It is particularly important to note that  $\sigma_z$ directly reflects the focusing capability of the imaging system in the elevational (axial) direction. Accordingly, we define the sensitivity map $h$ as the system's reversed impulse response function along the elevational direction: 
\begin{equation}
h(-z_c) = \exp\left(-\frac{z_c^2}{2\sigma_z^2}\right).
\end{equation}

This highly concise definition ensures that the Gaussian primitives maintain their Gaussian form throughout subsequent computations, thereby preserving the differentiability of the entire model and enabling highly efficient calculations. 
For practical imaging systems, $\sigma_z$ can be estimated either by theoretical or numerical methods (please refer to the supplementary for details), or it can be directly obtained through experimental calibration. 
For imaging data where such parameters are difficult to obtain, we assume the researcher used a scanning step $\delta_z$ that just satisfies the Nyquist criterion, in which case $\sigma_z \approx \delta_z$ .
Ablation study (Tab.~\ref{tab_ablation_study}) confirms the optimality of this choice for both 2D and 3D metrics.




\subsection{Gaussian Piling}
\label{pile}

We begin by considering a single Gaussian primitive. In this formulation, the rendered intensity $I$ arises from the convolution of the Gaussian primitive $g_c \sim \mathcal{G}(\boldsymbol{\mu}_c, \boldsymbol{\Sigma}_c)$  with the sensitivity map $h(z)$: 

\begin{equation}
I(x_c, y_c) = \left[ h(-z_c) * g_c(\mathbf{x}_c) \right]_{z_c = 0}.
\end{equation}


Expanding this convolution into its integral form yields:

\begin{equation}
\label{eq:conv_integ}
I(x_c, y_c, z_c = 0) = \int_{-\infty}^{\infty} h(t) \cdot g_c(x_c, y_c, t) \, dt.
\end{equation}

For detailed calculation process and fully expanded form, please refer to the supplementary materials. To achieve efficient computation, we phenomenologically decompose the process described by Eq. \ref{eq:conv_integ} into the following steps:


\noindent \textbf{Axial re-parameterization.} Finite slice thickness is encoded by injecting axial resolution term into the inverse covariance matrix in camera coordinates: :
\begin{equation}
\boldsymbol{\Sigma}_e^{-1}=\boldsymbol{\Sigma}_c^{-1}+\frac{\mathbf{e}_3\mathbf{e}_3^{\top}}{\sigma_z^2},\qquad
\boldsymbol{\mu}_e=\boldsymbol{\Sigma}_e\,\boldsymbol{\Sigma}_c^{-1}\boldsymbol{\mu}_c,
\end{equation}

\noindent where $\mathbf{e}_3=[0,0,1]^{\top}$. This operation shrinks the axial support without disturbing the lateral structure, producing an effective Gaussian  $f \sim \mathcal{G}(\boldsymbol{\mu}_e, \boldsymbol{\Sigma}_e)$ that matches the physics of finite slice thickness. 
We term this the \textbf {Focus Gaussian}. 
The term $\sigma_z$ here reconciles 2D slice fidelity with 3D volumetric consistency. 

\noindent \textbf{Opacity modulation.} To suppress contributions from out-of-focus primitives and stabilize reconstruction of thin slices, we modulate the opacity using the change in Mahalanobis distance induced by the axial injection:
\begin{equation}
\mathrm{opacity}_r = \exp\!\left(-\tfrac{1}{2}\big(\boldsymbol{\mu}_c^{\top}\boldsymbol{\Sigma}_c^{-1}\boldsymbol{\mu}_c-\boldsymbol{\mu}_e^{\top}\boldsymbol{\Sigma}_e^{-1}\boldsymbol{\mu}_e\big)\right).
\end{equation}

This factor has a clear physical interpretation: Gaussian primitives far from the focal vicinity of the slice undergo stronger transparency attenuation. This naturally down-weights primitives away from the slice and effectively prevents the accumulation of "ghosting" or floating artifacts across slices. Consequently, we cull primitives with opacity below a small threshold to reduce computational load. 


\noindent \textbf{Screen-space projection.} The 2D Gaussian parameters on the imaging plane are obtained by computing the marginal distribution of the Focus Gaussian's covariance matrix  $\boldsymbol{\Sigma}_e$ with respect to the $(x_c, y_c)$ plane: 
\begin{equation}
\boldsymbol{\Sigma}_{2d}=
\begin{bmatrix}
\boldsymbol{\Sigma}_e[0,0] & \boldsymbol{\Sigma}_e[1,0]\\
\boldsymbol{\Sigma}_e[1,0] & \boldsymbol{\Sigma}_e[1,1]
\end{bmatrix},\quad
\boldsymbol{\mu}_{2d}=[\boldsymbol{\mu}_e[0],\,\boldsymbol{\mu}_e[1]]^{\top}.
\end{equation}

It is important to note that the final rendered intensity $\tilde{\alpha}$ is normalized using the coefficient obtained from integration: 
\begin{equation}
\label{image_intensity}
\tilde{\alpha} = \alpha \cdot \mathrm{opacity}_r / \sqrt{\det(\boldsymbol{\Sigma}_{2d})}.
\end{equation}
The determinant term ensures brightness adapts to footprint size, preserving total energy as primitives stretch or shrink.

\noindent \textbf{Additive rasterization.} In slice imaging, pixel intensity originates from the integral of the Gaussian primitive opacities along the projection line, where contributions naturally superimpose without occlusion. We therefore employ additive accumulation: 

\begin{equation}
I({p}) = \sum_{i \in \mathcal{N}({p})} \tilde{\alpha}_i \exp\left(-\tfrac{1}{2}\mathbf{d}_i^{\top}\boldsymbol{\Sigma}_{2d,i}^{-1}\mathbf{d}_i\right),
\end{equation}
where $\mathcal{N}({p})$ denotes the  set of Gaussians overlapping pixel ${p}$, 
$\mathbf{d}_i$ is the vector from the 2D Gaussian center,  $\boldsymbol{\mu}_{2d}$, to the pixel,  ${p}$, and $\tilde{\alpha}_i$ 
is the rendered intensity from Eq.~\ref{image_intensity}. The gradient flow naturally follows 
the additive structure: $\frac{\partial I}{\partial \tilde{\alpha}_i} = \exp(\cdot)$, 
enabling efficient CUDA-accelerated optimization.

This rendering model is a principled design based on the inherent nature of volumetric slice imaging. Its advantages are primarily reflected in the following aspects:

Firstly, by coupling the intrinsic intensity information of the 3D volume data with the covariance of the 3D Gaussian primitives, 3D-to-2D consistency is enforced. As the same Gaussian primitive, observed by multiple slice images, inherently possesses the same set of covariance parameters, the continuity of the 3D structure is guaranteed. A Gaussian cannot exploit representation inconsistencies by fitting the image at the slice location while maintaining a different internal structure outside the focal plane, because the gradient information directly depends on the Focus Gaussian representation. This built-in regularization prevents the photometrically plausible but volumetrically inconsistent reconstructions observed in traditional 3DGS, where 2D projections look reasonable but the 3D structure collapses (Fig.~\ref{FIG_3d_abalation_conpare}).

Secondly, under the focus-aware model, the rendered slice is essentially the marginal distribution of an effective 3D Gaussian mixture model shaped by $\boldsymbol{\Sigma}_e$. This covariance-driven intensity representation completely discards the spherical harmonics (SH) coefficients introduced in standard 3DGS to fit the bidirectional reflectance distribution function (BRDF) model. This not only achieves natural compression (reducing the number of parameters per primitive by approximately 40\%) but also improves geometric fidelity.

\begin{figure}[!t]
\centerline{\includegraphics[width=\columnwidth]{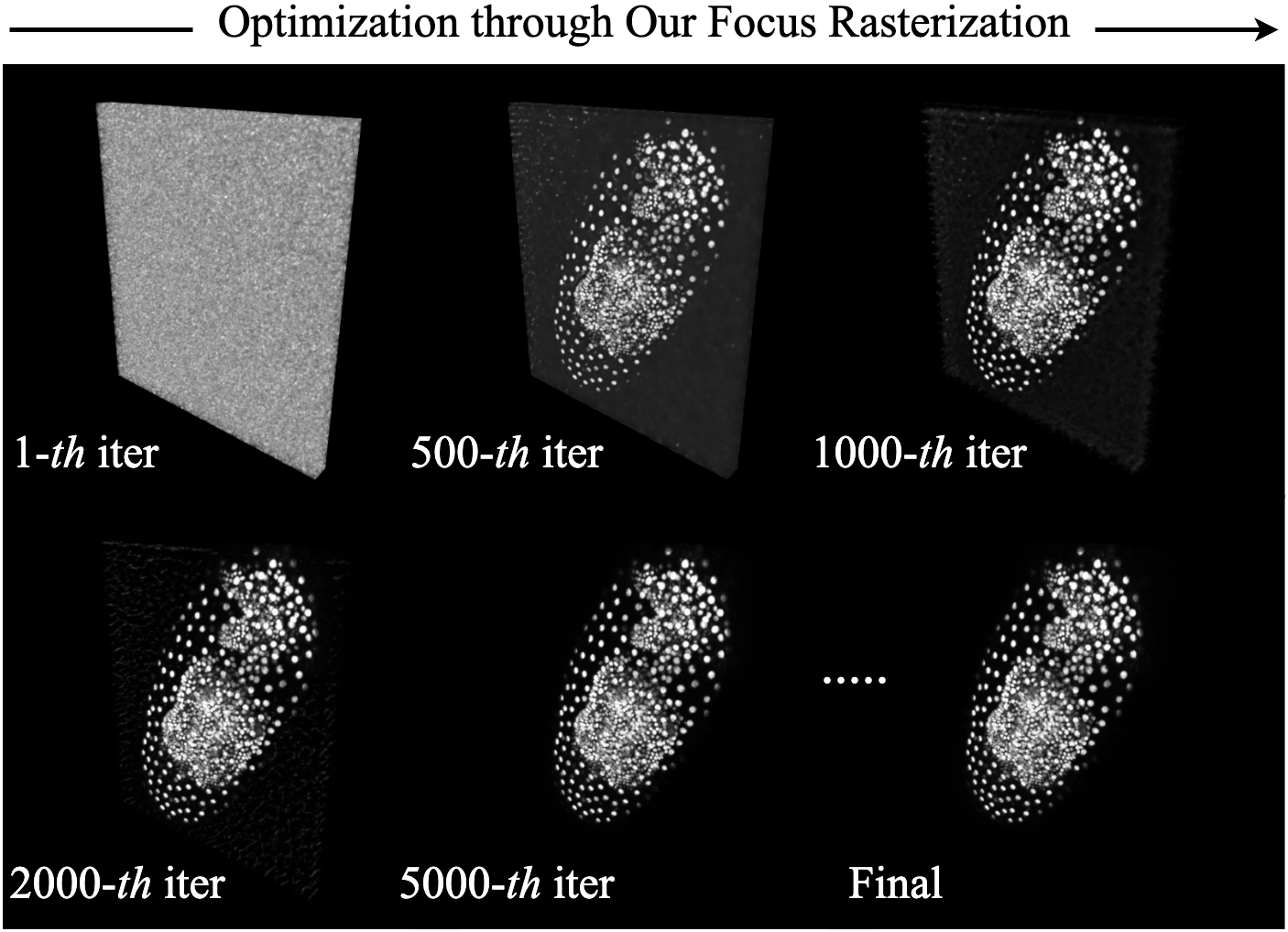}}
\caption{Reconstruction results of our GaussianPile at different iterations, visualized on the 3D data of a noisy \textit{Tribolium castaneum} embryo by confocal microscopy with high laser power, demonstrating high 3D fidelity throughout the optimization process.}
\label{FIG_3d_vis}
\end{figure}

\begin{figure*}[!ht]
\centerline{\includegraphics[width=17.4cm]{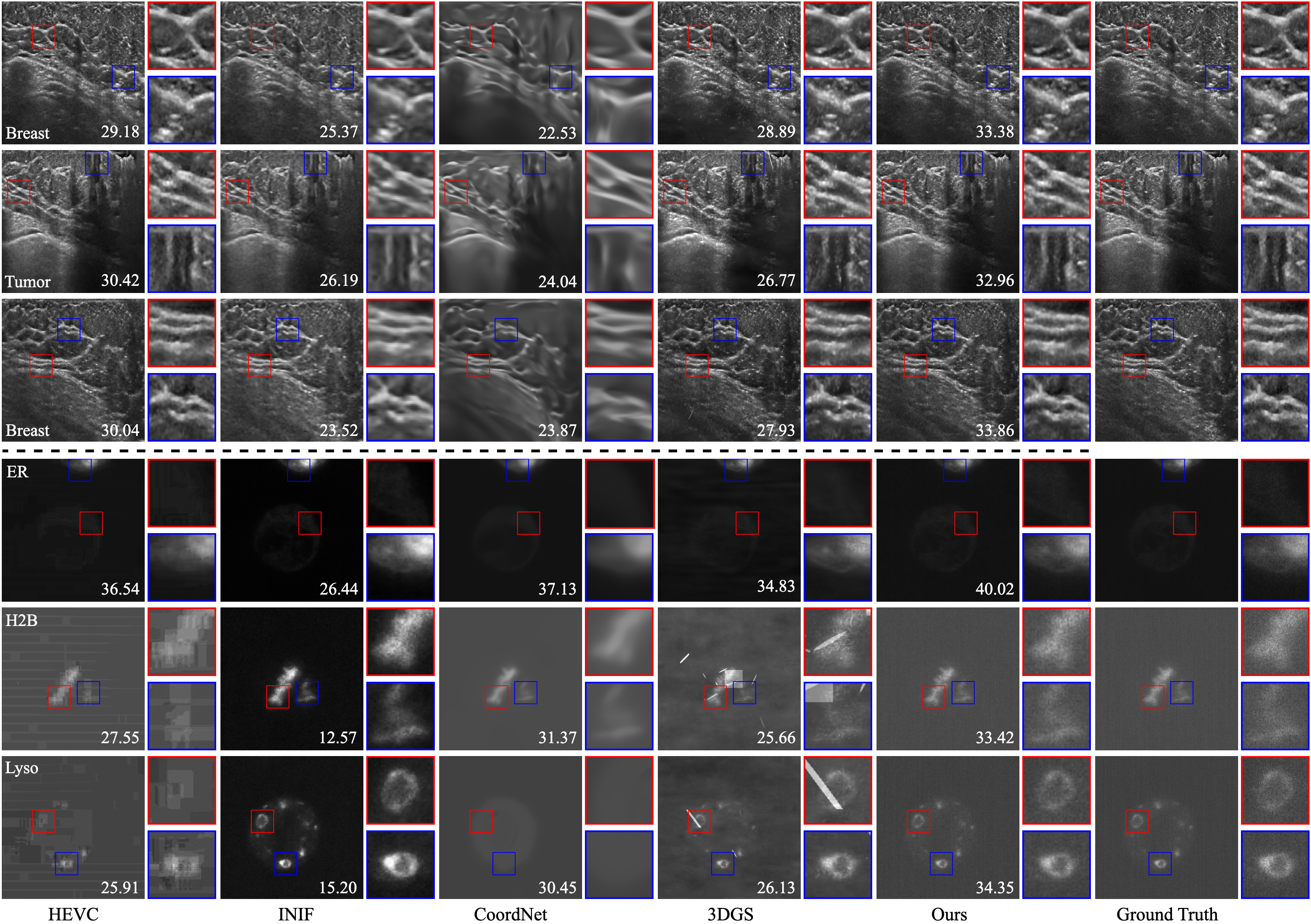}}
\vspace{-6pt}
\caption{Qualitative comparison of reconstruction results. Slice examples of different methods with PSNR (dB) shown at the bottom right of each image. The up three rows are from the TDSC-ABUS dataset \cite{luo2025tumor} and the down three rows are from the rDL-LSM dataset \cite{qiao2023rationalized}. Our method shows superior performance in modeling different regions (e.g. breast fibroglandular tissues and tumors) while preserving finer details compared to existing approaches.}
\label{FIG_quali_comparison}
\end{figure*}

\subsection{Optimization}
\label{optimize}

Training and saving follows the standard 3D Gaussian Splatting loop with several key adaptations. The optimization process is visualized in Fig.~\ref{FIG_3d_vis}.

\noindent \textbf{Backward pass.} We implement full gradient propagation through the focus-aware pipeline in CUDA. Given upstream gradients w.r.t. rendered pixels, the chain rule flows back through: (i) screen-space projection to effective Gaussian parameters ($\boldsymbol{\mu}_e$, $\boldsymbol{\Sigma}_e$), (ii) axial re-parameterization to camera-space ($\boldsymbol{\mu}_c$, $\boldsymbol{\Sigma}_c$), and (iii) camera transform to world-space learnable parameters ($\boldsymbol{\mu}$, $\mathbf{s}$, $\mathbf{q}$, $\alpha$). Special care is taken for the determinant term in $\tilde{\alpha} = \alpha \cdot \mathrm{opacity}_r / \sqrt{\det(\boldsymbol{\Sigma}_{2d})}$, whose gradient couples both opacity and covariance:
$$\frac{\partial \mathcal{L}}{\partial \alpha} = \frac{\partial \mathcal{L}}{\partial \tilde{\alpha}} \cdot \frac{\mathrm{opacity}_r}{\sqrt{\det(\boldsymbol{\Sigma}_{2d})}},
\quad
\frac{\partial \mathcal{L}}{\partial \boldsymbol{\Sigma}_e} \propto \frac{\partial \mathcal{L}}{\partial \tilde{\alpha}} \cdot \frac{\partial \tilde{\alpha}}{\partial \det} + \text{(conic)}.$$
Both kernels are CUDA-accelerated, maintaining real-time efficiency. Full derivations are provided in supplementary materials.

\noindent \textbf{Slice sampling.} At each iteration we randomly pick one virtual camera (i.e., one imaging slice) to render, ensuring coverage of the whole 3D image stack.

\noindent \textbf{Loss function.} Our framework is optimized using a photometric loss that supervises rendered slice projections. We adopt a weighted combination of L1 loss $\mathcal{L}_1$ and D-SSIM loss $\mathcal{L}_{ssim}$:
\begin{equation}
\mathcal{L} = \mathcal{L}_1 + \lambda \mathcal{L}_{ssim},
\end{equation}
where $\lambda$ balances perceptual and pixel-wise fidelity.

\noindent \textbf{Prune and densification.}
During training, we employ a densify-and-prune strategy based on tile-wise focus radii and Gaussian opacity scores. This adaptive mechanism progressively refines the Gaussian point cloud distribution, ensuring efficient structural coverage while preventing excessive axial overgrowth. After convergence, GaussianPile produces a differentiable point-cloud representation that reconstructs the input slices and provides improved apparent resolution along the scanning axis. The GaussianPile CUDA kernels serve all slice-based imaging modalities we tested, including automated 3D breast ultrasound (ABUS) and light-sheet microscopy (LSM), demonstrating the generality of the approach.

\noindent \textbf{Quantization and compression.} Unlike voxel grids, GaussianPile represents the scene as a sparse set of focus Gaussians, whose parameters are highly structured and spatially correlated. We exploit this property to design a tailored quantization and entropy-coding scheme. All Gaussians are first sorted in Morton (Z-order) space to expose local coherence. Positions are normalized to the scene bounding box and quantized with adaptive precision (typically 14 bits per axis); opacities are mapped to 12-bit integers in $[0,1]$; scales are log-transformed before 12-bit quantization to preserve multiplicative accuracy; and quaternions are confined to the positive half-sphere and quantized to 12 bits per component. Each attribute stream is delta-encoded and entropy-compressed using LZMA. This focus-aware encoding achieves compactness by leveraging the anisotropic sparsity and smooth parameter variation intrinsic to Gaussian splats, delivering $16\times$ reduction over voxel-based storage while retaining full 2D rendering fidelity and volumetric consistency.

\noindent \textbf{Differentiable Voxelization.} To evaluate 3D reconstruction quality and enable volumetric visualization, following \cite{zha2024r}, we integrate a differentiable voxelizer $\mathcal{V}$ that efficiently constructs an intensity volume $\mathbf{V} \in \mathbb{R}^{X \times Y \times Z}$ from world-space Gaussian primitives $\mathbb{G}^3 = \{g_i^3\}_{i=1,\ldots,M}$: $\mathbf{V} = \mathcal{V}(\mathbb{G}^3)$. The voxelizer partitions the 3D space into uniform $8^3$ tiles and culls Gaussians per tile, retaining only those whose spatial support (3-$\sigma$ ellipsoid) intersects the tile. Within each tile, voxel intensity at $\mathbf{x}$ is computed by aggregating the contributions of world-space Gaussians:
\begin{equation}
\sigma(\mathbf{x}) = \sum_{i=1}^{M} g_i^3(\mathbf{x}; \boldsymbol{\mu}_i, \boldsymbol{\Sigma}_i),
\end{equation}
where $g_i^3$ is the original Gaussian primitive in world coordinates before the re-parameterization in Sec.~\ref{pile}. Both forward and backward passes are CUDA-accelerated, achieving real-time query performance ($>100$ FPS). This enables efficient 3D structure evaluation and optional volumetric regularization during training.

\section{Experiments}
\subsection{Experimental Setting}

\begin{table*}[t]
  \centering
  \caption{2D quantitative comparison of GaussianPile and other methods.}
  \label{tab_2d_quanti_comparison}
  \vspace{-6pt}
  \resizebox{\textwidth}{!}{%
  \begin{tabular}{@{} l *{5}{c c c} @{}}
    \toprule
    \multirow{2}{*}{Method} 
      & \multicolumn{3}{c}{ABUS}
      & \multicolumn{3}{c}{rDL-LSM}
      & \multicolumn{3}{c}{TNNI1}
      & \multicolumn{3}{c}{hiPSC}
      & \multicolumn{3}{c}{Tribolium} \\
    \cmidrule(lr){2-4} \cmidrule(lr){5-7} \cmidrule(lr){8-10}
    \cmidrule(lr){11-13} \cmidrule(lr){14-16}
      & PSNR $^\uparrow$ & SSIM $^\uparrow$ & Time $^\downarrow$
      & PSNR $^\uparrow$ & SSIM $^\uparrow$ & Time $^\downarrow$
      & PSNR $^\uparrow$ & SSIM $^\uparrow$ & Time $^\downarrow$
      & PSNR $^\uparrow$ & SSIM $^\uparrow$ & Time $^\downarrow$
      & PSNR $^\uparrow$ & SSIM $^\uparrow$ & Time $^\downarrow$ \\
    \midrule
    HEVC~\cite{sze2014high}  & 29.67 & 0.729 & - & 29.34 & 0.615 & - & 35.76 & 0.863 & - & 19.73 & 0.450 & - & 35.51 & 0.710 & - \\
    INIF~\cite{dai2025implicit}  & 24.84 & 0.608 & 1h24m & 17.54 & 0.372 & 54m & 30.89 & 0.826 & 27m & 18.49 & 0.322 & 29m & 31.55 & 0.517 & 27m \\
    CoordNet~\cite{han2022coordnet}  & 23.32 & 0.517 & 60m & 31.98 & 0.631 & 47m & 30.12 & 0.827 & 32m & 24.78 & 0.627 & 60m & 29.42 & 0.687 & 46m \\
    NeurComp~\cite{lu2021compressive}  & 19.85 & 0.341 & 2h41m & 30.32 & 0.602 & 52m & 39.25 & 0.915 & 11m & 21.61 & 0.670 & 43m & 32.59 & 0.719 & 21m \\
    3DGS~\cite{kerbl20233d}  & 27.41 & 0.794 & 24m & 28.63 & 0.749 & 10m & 38.17 & 0.931 & 12m & 20.15 & 0.522 & 26m & 27.81 & 0.715 & 1h4m \\
    \midrule
    Ours (iter=10k)    & 32.25  & 0.803 & \textbf{4m}
                       & 33.78  & 0.783 & \textbf{3m} 
                       & 40.87  & 0.943 & \textbf{2m}
                       & 32.03  & 0.709 & \textbf{2m}
                       & 35.32  & 0.820 & \textbf{3m} \\
    Ours (iter=30k)    & \textbf{33.07} & \textbf{0.825} & 13m                                 & \textbf{34.57} & \textbf{0.791} & 9m 
                       & \textbf{42.08} & \textbf{0.948} & 5m 
                       & \textbf{32.59} & \textbf{0.728} & 8m
                       & \textbf{36.14} & \textbf{0.848} & 7m \\
    \bottomrule
  \end{tabular}%
  }
  \vspace{-4pt}
\end{table*}

\begin{table*}[t]
  \centering
  \caption{Compression comparison of GaussianPile and other methods. Memory size is reported in megabytes (MB).}
  \label{tab_Comp_comparison}
  \vspace{-6pt}
  \resizebox{\textwidth}{!}{%
  \begin{tabular}{@{} l *{7}{c c} @{}}
    \toprule
    \multirow{2}{*}{Method} 
      & \multicolumn{2}{c}{ABUS}
      & \multicolumn{2}{c}{rDL-LSM}
      & \multicolumn{2}{c}{TNNI1}
      & \multicolumn{2}{c}{hiPSC}
      & \multicolumn{2}{c}{Tribolium high}
      & \multicolumn{2}{c}{Tribolium low}
      & \multicolumn{2}{c}{Tribolium very low} \\
    \cmidrule(lr){2-3} \cmidrule(lr){4-5} \cmidrule(lr){6-7}
    \cmidrule(lr){8-9} \cmidrule(lr){10-11} \cmidrule(lr){12-13} \cmidrule(lr){14-15}
      & Mem.$^\downarrow$ & Comp.$^\uparrow$
      & Mem.$^\downarrow$ & Comp.$^\uparrow$
      & Mem.$^\downarrow$ & Comp.$^\uparrow$
      & Mem.$^\downarrow$ & Comp.$^\uparrow$
      & Mem.$^\downarrow$ & Comp.$^\uparrow$
      & Mem.$^\downarrow$ & Comp.$^\uparrow$
      & Mem.$^\downarrow$ & Comp.$^\uparrow$ \\
    \midrule
    INIF~\cite{dai2025implicit}  & 7.8 & 15$\times$ & 4.9 & 15$\times$ & \textbf{1.8} & \textbf{18$\times$} & 4.7 & 15$\times$ & 3.1 & 16$\times$ & 3.1 & 16$\times$ & 3.1 & 16$\times$ \\
    CoordNet~\cite{han2022coordnet}  & 10.7 & 11$\times$ & 5.2 & 14$\times$ & 3.2 & 10$\times$ & 5.5 & 13$\times$ & 5.6 & 9$\times$ & 5.6 & 9$\times$ & 5.6 & 9$\times$ \\
    NeurComp~\cite{lu2021compressive}  & 7.4 & 16$\times$ & \textbf{4.5} & \textbf{16$\times$} & 2.0 & 16$\times$ & 4.5 & 16$\times$ & 3.0 & 16$\times$ & 3.0 & 16$\times$ & 3.0 & 16$\times$ \\
    3DGS~\cite{kerbl20233d}  & 691.7 & 0.2$\times$ & 50.2 & 1.5$\times$ & 10.6 & 3$\times$ & 173.6 & 0.4$\times$ & 28.2 & 1.7$\times$ & 378.4 & 0.1$\times$ & 45.0 & 1.1$\times$ \\
    \midrule
    Ours w/o Quant.    & 49.8 & 2$\times$
                       & 22.6 & 3$\times$
                       & 9.7 & 3$\times$
                       & 21.8 & 3$\times$
                       & 9.3 & 5$\times$
                       & 11.3 & 4$\times$ & 11.0 & 4$\times$ \\
    Ours               & \textbf{6.2}  & \textbf{19$\times$}
                       & \textbf{4.5}  & \textbf{16$\times$}
                       & 2.0  & 16$\times$
                       & \textbf{4.2}  & \textbf{17$\times$}
                       & \textbf{1.8}  & \textbf{26$\times$}
                       & \textbf{2.4}  & \textbf{20$\times$} 
                       & \textbf{2.3}  & \textbf{20$\times$} \\
    \bottomrule
  \end{tabular}%
  }
  \vspace{-4pt}
\end{table*}

\noindent \textbf{Dataset.} We evaluate our method on several public real-world datasets. Our experimental setup includes: (i) 30 cases from the TDSC-ABUS dataset \cite{luo2025tumor}; (ii) 30 samples from the rDL-LSM dataset \cite{qiao2023rationalized}, comprising 10 samples each from the three distinct channels (ER, H2B, and Lyso); and (iii) 8 samples across three cellular datasets: Noisy Tribolium data \cite{weigert2018content} with three levels of laser power (high, low and very low), hiPSC cell data \cite{viana2023integrated}, and TNNI1-stained cell data \cite{chen2018allen}.

\noindent \textbf{Implementation details.} GaussianPile is implemented in PyTorch \cite{paszke2019pytorch} and CUDA \cite{sanders2010cuda}, and trained with the Adam optimizer \cite{kinga2015method} for 30k iterations. Learning rates for position, opacity, scale, and rotation are initially set as 0.0006, 0.02, 0.002, and 0.001, respectively, and exponentially to 0.1 of their initial values. We initialize $M = 1000\text{k}$ Gaussians for all datasets with a opacity minimal threshold $\tau = 0.02$. Adaptive control runs from $500$ to $25000$ iterations with a gradient threshold of $0.00005$. $\lambda$ is set to $0.2$. All experiments run on NVIDIA A800 GPUs. We employ four metrics for comprehensive evaluation: 2D PSNR and 2D SSIM measure the quality between rendered slices and ground truth slices, while 3D PSNR and 3D SSIM \cite{wang2004image} assess the fidelity between our voxelized volume and the ground truth volume. Beyond reconstruction quality, we report running time, memory size, and compression ratio to evaluate efficiency and compression performance.

\subsection{Results and Evaluation}
We compare GaussianPile with traditional method (HEVC \cite{sze2014high}), SOTA INR-based baselines (INIF \cite{dai2025implicit}, CoordNet \cite{han2022coordnet}, and NeurComp \cite{lu2021compressive}), and slice-adapted original 3DGS \cite{kerbl20233d}. Tabs.~\ref{tab_2d_quanti_comparison} and \ref{tab_3d_quanti_comparison} report the quantitative results in both 2D slice and 3D volume. Note that we do not report the running time for HEVC as it is almost instant. GaussianPile achieves the best performance across all ultrasound datasets and most microscopy datasets, consistently outperforming the INR-based baselines while also yielding clearly better 3D fidelity than slice-adapted 3DGS. In terms of efficiency, our method converges to high-quality results in $8$ minutes on average, which is about $5\times$ faster than INIF. Fig.~\ref{FIG_quali_comparison} shows representative visual comparisons. Due to space limitations, we include qualitative results for INIF and CoordNet as representative INR baselines. HEVC tends to blur fine structural details, INR methods often lose high-frequency information, and 3DGS suffers from floating artifacts because it lacks an appropriate slice formation model. In contrast, GaussianPile better recovers sharp structures, such as tumors in breast ultrasound and fine intracellular details in microscopy data, while preserving smooth homogeneous regions. Tab.~\ref{tab_Comp_comparison} further compares memory size and compression ratio. We do not include HEVC in this comparison because its extreme compression ratios come at the expense of substantial fidelity loss, which is inconsistent with our goal of high-fidelity volumetric reconstruction. Overall, GaussianPile achieves the best balance between quality, efficiency, and compression performance.

\subsection{Ablation Study}

\begin{table}[!t]
  \centering
  \scriptsize
  \setlength{\tabcolsep}{1.2pt}
  \renewcommand{\arraystretch}{1.0}
  \caption{3D quantitative results of baselines and our method.}
  \label{tab_3d_quanti_comparison}
  \vspace{-6pt}
  \begin{tabular}{@{} l *{2}{c c c c c c} @{}}
    \toprule
    \multirow{2}{*}{Method} 
      & \multicolumn{2}{c}{ABUS}
      & \multicolumn{2}{c}{rDL-LSM}
      & \multicolumn{2}{c}{TNNI1}
      & \multicolumn{2}{c}{hiPSC} 
      & \multicolumn{2}{c}{Tribolum} \\
    \cmidrule(lr){2-3} \cmidrule(lr){4-5} \cmidrule(lr){6-7} \cmidrule(lr){8-9} \cmidrule(lr){10-11}
      & PSNR & SSIM
      & PSNR & SSIM 
      & PSNR & SSIM 
      & PSNR & SSIM 
      & PSNR & SSIM \\
    \midrule
    CoordNet~\cite{han2022coordnet}  & 18.75 & 0.454 & 32.03 & 0.689 & 29.26 & 0.716 & 22.23 & 0.642 & 28.56 & 0.693  \\
    NeurComp~\cite{lu2021compressive}  & 19.38 & 0.248 & 31.08 & 0.586 & 39.53 & 0.940 & 22.54 & 0.653 & 32.51 & 0.667  \\
    3DGS~\cite{kerbl20233d}  & 28.49 & 0.828 & 25.39 & 0.665 & 26.75 & 0.780 & 15.20 & 0.349 & 21.46 & 0.583 \\
    Ours  & \textbf{33.22} & \textbf{0.876} & \textbf{34.47} & \textbf{0.814} & \textbf{40.17} & \textbf{0.952} & \textbf{33.73} & \textbf{0.752} & \textbf{35.05} & \textbf{0.860} \\
    \bottomrule
  \end{tabular}%
  \vspace{-6pt}
\end{table}


\begin{figure}[!t]
\centerline{\includegraphics[width=\columnwidth]{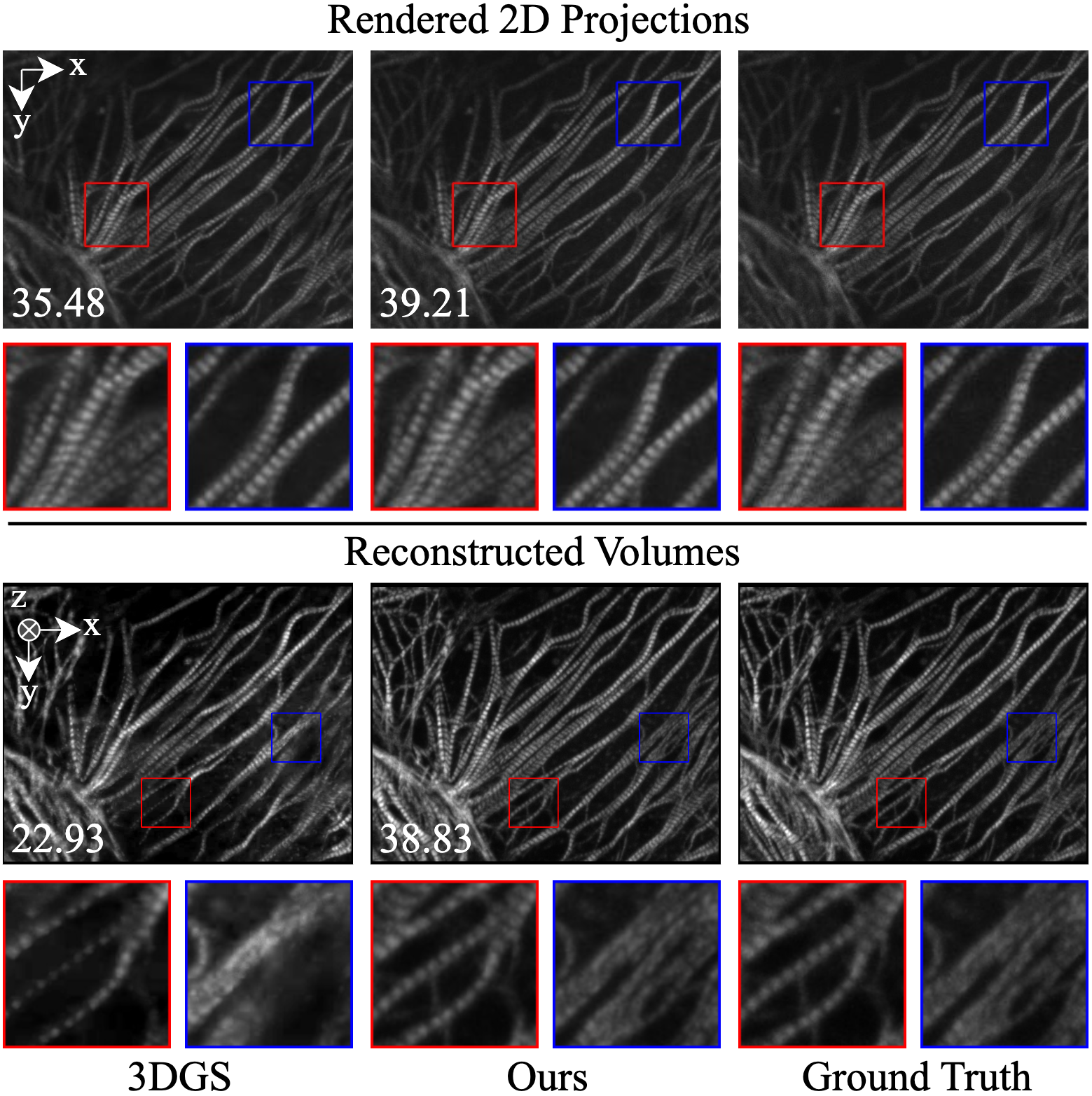}}
\vspace{-2pt}
\caption{Results of 3DGS and our method with PSNR (dB) indicated on each image. We show slices of 3DGS queried from elevational angle (z-axis). Although 3DGS can produce plausible 2D projections, its reconstructed volume exhibits poor quality.}
\label{FIG_3d_abalation_conpare}
\end{figure}

\noindent \textbf{Focus forward model.} To demonstrate the impact of  focus forward model and slice thickness discussed in early Sec.~\ref{modeling}, we develop a slice-adapted version of 3DGS, inspired by FruitNinja \cite{wu2025fruitninja}. Tab.~\ref{tab_2d_quanti_comparison} and Tab.~\ref{tab_3d_quanti_comparison} show that integrating the physical focus model benefits both 2D rendering and 3D reconstruction. Fig.~\ref{FIG_3d_abalation_conpare} visualizes rendering and reconstruction results. While 3DGS renders reasonable 2D projections, its reconstruction quality is significantly worse than ours. The conflicting 2D and 3D performances indicate that 3DGS, despite fitting slice images well, does not accurately model the 3D internal structure. In contrast, our method shows superior 3D fidelity. We further discuss the influence of slice thickness $\sigma_z$. As shown in Tab.~\ref{tab_ablation_study}, our model performs best when setting slice thickness equals to real pixel spacing along scanning view, denoted as $\delta_z$, suggesting accuracy of our physical focus model.

\noindent \textbf{Quantization.} Ablation studies across all datasets evaluate the impact of the Morton Gaussian quantizer (Quant.) on compression. The original voxel sizes of ABUS, rDL-LSM, TNNI1, hiPSC, and Tribolium datasets are $118, 75.5, 33.3, 71.5$ and $47$ MB, respectively. As Tab.~\ref{tab_Comp_comparison} shows, while our baseline yields a $3.4\times$ average compression ratio, adding the quantizer boosts it to $19\times$ without degrading reconstruction quality.

\begin{table}[!t]
  \centering
  \setlength{\tabcolsep}{6pt}
  \caption{Ablation results with our choices in bold.}
  \label{tab_ablation_study}
  \vspace{-5pt}
  \begin{tabular}{@{} l c c c c @{}}
    \toprule
    Tribolium high & PSNR $^\uparrow$ & SSIM $^\uparrow$ & Time $^\downarrow$ & Gau. \\
    \midrule
    $\sigma_z=\delta_z/10$  & 34.25 & 0.927 & 8m9s & 215k \\
    $\sigma_z=\delta_z/2$  & 36.44 & 0.942 & 8m18s & 214k \\
    $\bm{\sigma_z=\delta_z}$  & \textbf{36.67} & \textbf{0.944} & \textbf{7m23s} & 211k \\
    $\sigma_z=2\delta_z$  & 36.12 & 0.940 & 8m8s & 162k \\
    $\sigma_z=10\delta_z$  & 23.05 & 0.858 & 9m17s & 66k \\
    \midrule
    $M=50\text{k}$  & 34.12 & 0.932 & 7m28s & 85k \\
    $M=200\text{k}$  & 34.33 & 0.935 & 7m26s & 87k \\
    $\bm{M=1000\textbf{k}}$  & 36.67 & 0.944 & \textbf{7m23s} & 211k \\
    $M=1500\text{k}$  & 36.78 & 0.944 & 8m2s& 235k \\
    $M=2000\text{k}$  & \textbf{36.82} & \textbf{0.945} & 8m27s & 272k \\
    \midrule
    Init.~=~Grid  & 36.30 & 0.942 & 9m4s & 220k \\
    \textbf{Init.~=~Random}  & \textbf{36.67} & \textbf{0.944} & \textbf{7m23s} & 211k \\
    \bottomrule
  \end{tabular}%
  \vspace{-8pt}
\end{table}

\noindent \textbf{Component analysis.} The initialization ablation (last block of Tab.~\ref{tab_ablation_study}) shows that while structured grids seed Gaussians uniformly in 3D space to ensure even spatial coverage, they yield larger models, longer training, and poorer metrics. In contrast, random initialization, implemented with stochastic spatial and opacity perturbations, naturally guides densification toward relevant regions, resulting in fewer primitives, faster convergence, and higher fidelity.

\noindent \textbf{Parameter analysis.} We perform parameter analysis on the number of initialized Gaussians $M$. The results are shown in the second block of Tab.~\ref{tab_ablation_study}, our GaussianPile achieves quality-efficiency balance at $1000$k initialized Gaussians.





\section{Discussion and Conclusion}
\vspace{-3pt}
In this work, we introduced GaussianPile, a novel representation that adapts 3D Gaussian Splatting to slice-based volumetric imaging. 
Our core contribution is an explicit, physics-based focus-aware forward model that accounts for the finite slice thickness of real imaging systems. 
This formulation enables anisotropic 3D Gaussians to faithfully capture not only surface-like features but also the complete internal structure of a specimen—a capability absent in standard 3DGS.

GaussianPile preserves the key strengths of Gaussian-based representations: high efficiency and inherent compressibility. 
With our optimized CUDA implementation, the model converges to high-quality results in minutes rather than hours. The structured sparsity of the Gaussian representation, combined with a tailored quantization and entropy coding scheme, achieves compression ratios of up to 26×, offering a practical solution for managing large-scale volumetric datasets. Beyond PSNR/SSIM, downstream segmentation results in supplementary also show improved Dice/IoU and fewer boundary errors on reconstructions, indicating that the method better preserves task-relevant structural fidelity.

Despite these strengths, GaussianPile has certain limitations. 
The current model assumes a spatially invariant PSF, while real optical systems often exhibit spatially varying aberrations. 
A promising future direction is to incorporate a learnable, spatially varying PSF to improve reconstruction accuracy \cite{yang2025image}. Nevertheless, generality experiments in supplementary suggest that GaussianPile remains stable under moderate PSF mismatch and shows promising generalization to diverse anisotropic slice-based modalities, although accommodating such deviations may require more Gaussian primitives and thus reduce compression efficiency.
In addition, the method remains sensitive to severely noisy or undersampled data, where optimization may lead to over-smoothing. 
Integrating semantic or physical priors could enhance robustness in such challenging scenarios \cite{zhu2025sparse}. 
Going further, learning a generalized prior from large-scale volumetric data could enable one-shot, feed-forward reconstruction, bypassing costly iterative optimization and accelerating inference for real-time use \cite{li2024learning}. 
Finally, extending GaussianPile to model 4D spatiotemporal data—such as in live-cell imaging—would greatly broaden its applicability to dynamic biological processes \cite{yang2024deformable}.

Notwithstanding these opportunities for extension, GaussianPile provides a practical and powerful framework for handling the growing data volumes produced by modern biomedical imaging, thereby establishing a solid foundation for future advances in volumetric representation and compression.
\section{Acknowledgment}
This paper is supported by National Key R\&D Program of China, under Grant 2024YFC2421800; National Natural Science Foundation of China, under Grant 62505153 and 62475129; China Postdoctoral Science Foundation, under Grant 2024M761726; Zhongguancun Academy, under Grant 20240311; Xishan-Tsinghua Industry-Acdemia-Research Integration Initiative, under Grant 20242002209; the Fuzhou Institute for Data Technology.
{
    \small
    \bibliographystyle{ieeenat_fullname}
    \bibliography{main}
}

\clearpage
\setcounter{page}{1}
\maketitlesupplementary

\section{Derivation of Forward rendering process}
\label{sec:derivation_of_fgf}

We will analytically construct the computational pathway from the imaging target composed of 3D Gaussian primitives $g(\mathbf{x})$ with position $\boldsymbol{\mu} \in \mathbb{R}^3$, covariance $\boldsymbol{\Sigma}$, and density $a$ to the final imaging output. Specifically, each primitive can be represented by an anisotropic 3D Gaussian:
\begin{equation}
g(\mathbf{x}) = \alpha \cdot \exp\left(-\frac{1}{2}(\mathbf{x}-\boldsymbol{\mu})^{\top}\boldsymbol{\Sigma}^{-1}(\mathbf{x}-\boldsymbol{\mu})\right),
\end{equation}
where $\mathbf{x}=[x,y,z]^{\top}$ represents the world coordinates. To avoid information redundancy, each Gaussian primitive in the computation contains the following parameters $G=(\boldsymbol{\mu}, \bm{s}, \bm{q}, \alpha)$, which include: the 3D position $\boldsymbol{\mu}$, the anisotropic scale $\bm{s}\in\mathbb{R}^3$ controlling the anisotropic scaling, the orientation quaternion $\bm{q}\in\mathbb{H}$ representing spatial orientation, and the intensity coefficient or opacity $\alpha$.

During the CUDA rasterization process, to enable differentiable rendering, we need to convert the scale $\bm{s}$ and orientation $\bm{q}$ into a 3D covariance matrix $\boldsymbol{\Sigma}$ that describes the Gaussian shape. This conversion follows the classical Gaussian splatting pipeline: first, obtain the diagonal scaling matrix considering the scaling factor through $\bm{S}=\mathrm{diag}(\mathrm{mod}\cdot\bm{s})$; second, convert the quaternion to a rotation matrix via $\bm{R}=\mathrm{quatToMatrix}(\bm{q})$; finally, synthesize the covariance matrix in the world coordinate system through the transformation $\boldsymbol{\Sigma}=\bm{R}\bm{S}\bm{S}^{\top}\bm{R}^{\top}$.

We conduct our analysis in the camera coordinate system, thus it is first necessary to transform the Gaussian primitives from the world coordinate system to the camera coordinate system. For each virtual slice, we utilize its exported view matrix $[R_c | \mathbf{t}]$ to transform the Gaussian center $\boldsymbol{\mu}$ and covariance matrix $\boldsymbol{\Sigma}$ from the world coordinate system to the camera coordinate system (focused imaging plane). The transformation formulas are as follows:
\begin{equation}
\boldsymbol{\mu}_c = R_c \cdot \boldsymbol{\mu} + \mathbf{t},
\end{equation}
\begin{equation}
\boldsymbol{\Sigma}_c = R_c \cdot \boldsymbol{\Sigma} \cdot R_c^{\top},
\end{equation}
where $R_c$ is a $3 \times 3$ rotation matrix and $\mathbf{t}$ is a $3 \times 1$ translation vector.

We assume the point spread function (PSF) of the imaging system is spatially invariant and can be represented by a 3D Gaussian function:
\begin{equation}
\text{psf}(\mathbf{x}_c) = \frac{1}{(2\pi)^{3/2} \sigma_x \sigma_y \sigma_z} \exp\left(-\frac{1}{2} \left( \frac{x_c^2}{\sigma_x^2} + \frac{y_c^2}{\sigma_y^2} + \frac{z_c^2}{\sigma_z^2} \right) \right),
\end{equation}
where $\sigma_x$, $\sigma_y$, and $\sigma_z$ represent the spatial resolution of the imaging system in the $x$, $y$, and $z$ directions, respectively. Notably, $\sigma_z$ directly reflects the focusing capability of the imaging system in the elevational (axial) direction. Based on this, we define the sensitivity map function $h(z)$ as the reversed system's axial impulse response:
\begin{equation}
h(-z_c) = \exp\left( -\frac{z_c^2}{2\sigma_z^2} \right).
\end{equation}

Generally, $\sigma_z$ relates to the numerical aperture ${NA}_{ele}$ in the elevational direction and the imaging wavelength $\lambda$, and can be calculated using the formula:
\begin{equation}
\sigma_z \approx 0.27 \cdot \frac{\lambda}{{NA}_{ele}}.
\end{equation}
For specific imaging systems, this formula can be further refined. For instance, for a linear optical imaging system with a numerical aperture ${NA}$:
\begin{equation}
\sigma_z \approx \frac{\lambda}{{NA}^2}.
\end{equation}
For an ultrasound imaging system:
\begin{equation}
\sigma_z \approx k \frac{\lambda F}{D_{ele}},
\end{equation}
where $\lambda$ is the ultrasonic wavelength, $F$ is the transducer focal length, and $D_{ele}$ is the effective aperture size of the ultrasonic transducer in the elevational direction.

According to the above formulation, for a single Gaussian primitive, the rendered image $I$ can be obtained by the convolution of the Gaussian primitive $g_c(\mathbf{x}_c)$ in the camera coordinates with the reversed sensitivity map function $h(-z)$, evaluated at $z_c=0$:
\begin{equation}
I(x_c, y_c) = \left[ h(-z) * g_c(\mathbf{x}_c) \right] \big|_{z_c=0}.
\end{equation}
Expanding the above equation into its integral form yields:
\begin{equation}
I(x_c, y_c, z_c=0) = \int_{-\infty}^{\infty} h(t) \cdot g_c(x_c, y_c, t)  dt.
\end{equation}

Since the integrand \( g_e(\mathbf{x}) = h(\mathbf{x}) g_c(\mathbf{x}) \) is the product of two Gaussian functions, it remains Gaussian in form. We refer to this effective Gaussian as the focus Gaussian.

When deriving the parameters of the focus Gaussian, we consider the sum of the exponent terms of the two Gaussian functions. The exponent of the Gaussian \( g_c(\mathbf{x}) \) in camera coordinates is:
\begin{equation}
-\frac{1}{2} (\mathbf{x} - \boldsymbol{\mu}_c)^\top \boldsymbol{\Sigma}_c^{-1} (\mathbf{x} - \boldsymbol{\mu}_c).
\end{equation}
The exponent of the sensitivity map function \( h(\mathbf{x}) \) is:
\begin{equation}
-\frac{z_c^2}{2\sigma_z^2} = -\frac{1}{2\sigma_z^2} \mathbf{x}^\top (\mathbf{e}_3 \mathbf{e}_3^\top) \mathbf{x},
\end{equation}
where \( \mathbf{e}_3 = [0, 0, 1]^\top \). Therefore, the total exponent is:
\begin{equation}
\begin{split}
-\frac{1}{2} \left[ (\mathbf{x} - \boldsymbol{\mu}_c)^\top \boldsymbol{\Sigma}_c^{-1} (\mathbf{x} -  \boldsymbol{\mu}_c) + \frac{1}{\sigma_z^2} \mathbf{x}^\top (\mathbf{e}_3 \mathbf{e}_3^\top) \mathbf{x} \right].
\end{split}
\end{equation}

Expanding the first term:
\begin{equation}
\begin{split}
&(\mathbf{x} - \boldsymbol{\mu}_c)^\top \boldsymbol{\Sigma}_c^{-1} (\mathbf{x} - \boldsymbol{\mu}_c) = \\
&\mathbf{x}^\top \boldsymbol{\Sigma}_c^{-1} \mathbf{x} - 
2\mathbf{x}^\top \boldsymbol{\Sigma}_c^{-1} \boldsymbol{\mu}_c + \boldsymbol{\mu}_c^\top \boldsymbol{\Sigma}_c^{-1} \boldsymbol{\mu}_c.
\end{split}
\end{equation}

Thus the complete exponent becomes:
\begin{equation}
\begin{split}
-\frac{1}{2} \bigg[ \mathbf{x}^\top \left( \boldsymbol{\Sigma}_c^{-1} + \frac{\mathbf{e}_3 \mathbf{e}_3^\top}{\sigma_z^2} \right) \mathbf{x} \\
 - 2\mathbf{x}^\top \boldsymbol{\Sigma}_c^{-1} \boldsymbol{\mu}_c \
 + \boldsymbol{\mu}_c^\top \boldsymbol{\Sigma}_c^{-1} \boldsymbol{\mu}_c \bigg].
\end{split}
\end{equation}

Comparing this with the standard Gaussian form \( -\frac{1}{2} (\mathbf{x} - \boldsymbol{\mu}_e)^\top \boldsymbol{\Sigma}_e^{-1} (\mathbf{x} - \boldsymbol{\mu}_e) \), we can identify:
\begin{equation}
\boldsymbol{\Sigma}_e^{-1} = \boldsymbol{\Sigma}_c^{-1} + \frac{\mathbf{e}_3 \mathbf{e}_3^\top}{\sigma_z^2},
\end{equation}
and
\begin{equation}
\boldsymbol{\Sigma}_e^{-1} \boldsymbol{\mu}_e = \boldsymbol{\Sigma}_c^{-1} \boldsymbol{\mu}_c.
\end{equation}

Solving for \( \boldsymbol{\mu}_e \):
\begin{equation}
\boldsymbol{\mu}_e = \boldsymbol{\Sigma}_e \boldsymbol{\Sigma}_c^{-1} \boldsymbol{\mu}_c.
\end{equation}

The remaining constant term \( \boldsymbol{\mu}_c^\top \boldsymbol{\Sigma}_c^{-1} \boldsymbol{\mu}_c - \boldsymbol{\mu}_e^\top \boldsymbol{\Sigma}_e^{-1} \boldsymbol{\mu}_e \) is absorbed into the opacity modulation term:
\begin{equation}
\text{opacity}_r = \exp\left( -\frac{1}{2} \left( \boldsymbol{\mu}_c^\top \boldsymbol{\Sigma}_c^{-1} \boldsymbol{\mu}_c - \boldsymbol{\mu}_e^\top \boldsymbol{\Sigma}_e^{-1} \boldsymbol{\mu}_e \right) \right).
\end{equation}

Finally, the rendered image \( I(x_c, y_c) \) can be expressed as:
\begin{equation}
\begin{split}
&I(x_c, y_c) = \alpha \cdot \text{opacity}_r \\ &\int_{-\infty}^{\infty} \exp\left( -\frac{1}{2} (\mathbf{x} - \boldsymbol{\mu}_e)^\top \boldsymbol{\Sigma}_e^{-1} (\mathbf{x} - \boldsymbol{\mu}_e) \right)\quad dz_c.
\end{split}
\end{equation}

The integrand is clearly a three-dimensional Gaussian distribution. Integrating over \( z_c \) yields the marginal distribution with respect to \( x_c \) and \( y_c \), which is a two-dimensional Gaussian distribution. Its mean and covariance matrix are determined by the corresponding components of the original mean vector and covariance matrix. Based on this, we directly express the computed result as:
\begin{equation}
\begin{split}
&I(x_c, y_c) =  \alpha \cdot \frac{\text{opacity}_r}{\sqrt{|\boldsymbol{\Sigma}_{2d}|}} \\
& \cdot \exp\left( -\frac{1}{2} ([x_c, y_c]^{\top} - \boldsymbol{\mu}_{2d})^{\top} \boldsymbol{\Sigma}_{2d}^{-1} ([x_c, y_c]^{\top} - \boldsymbol{\mu}_{2d}) \right),
\end{split}
\end{equation}
where \( \boldsymbol{\Sigma}_{2d} \) and \( \boldsymbol{\mu}_{2d} \) are the components of \( \boldsymbol{\Sigma}_e \) and \( \boldsymbol{\mu}_e \) along the \( x \) and \( y \) directions, respectively:
\begin{equation}
\boldsymbol{\Sigma}_{2d} = \begin{bmatrix}
\boldsymbol{\Sigma}_e[0,0] & \boldsymbol{\Sigma}_e[1,0] \\
\boldsymbol{\Sigma}_e[1,0] & \boldsymbol{\Sigma}_e[1,1]
\end{bmatrix}, \quad \boldsymbol{\mu}_{2d} = [\boldsymbol{\mu}_e[0], \boldsymbol{\mu}_e[1]]^{\top}.
\end{equation}

To facilitate the rendering of the 2D image, we transform the above equation into the screen space. For a pixel \( p \), we use \( \mathbf{d} \) to denote the 2D vector from the 2D Gaussian center \( \boldsymbol{\mu}_{2d} \) to the coordinates of pixel \( p \). The rendered image can then be expressed as:
\begin{equation}
I(p) = \tilde{\alpha} \cdot \exp\left( -\frac{1}{2} \mathbf{d}^{\top} \boldsymbol{\Sigma}_{2d}^{-1} \mathbf{d} \right),
\end{equation}
where \( \tilde{\alpha} \) represents the modulated final rendering intensity:
\begin{equation}
\tilde{\alpha} = \alpha \cdot \frac{\text{opacity}_r}{\sqrt{|\boldsymbol{\Sigma}_{2d}|}}.
\end{equation}

Finally, for scenarios involving multiple Gaussian primitives, due to the linearity of the imaging model, we only need to sum over all Gaussian primitives:
\begin{equation}
I(p) = \sum_{i \in \mathcal{N}(p)} \tilde{\alpha}_i \cdot \exp\left( -\frac{1}{2} \mathbf{d}_i^{\top} \boldsymbol{\Sigma}_{2d,i}^{-1} \mathbf{d}_i \right),
\end{equation}
where \( \mathcal{N}(p) \) denotes the set of Gaussians overlapping with pixel \( p \).

\section{Derivation of Differentiable Backward Pass}

\subsection{Forward Process Review}
\noindent \textbf{2D mean:}
\begin{equation}
    \boldsymbol{\mu}_{2d} = [\boldsymbol{\mu}_e[0], \boldsymbol{\mu}_e[1]]^{\top}
\end{equation}

\noindent \textbf{2D inverse covariance matrix:}
\begin{equation}
    \boldsymbol{\Sigma}_{2d}^{-1} = \frac{1}{\det (\boldsymbol{\Sigma}_{2d})} 
    \begin{bmatrix}
    \boldsymbol{\Sigma}_e[1,1] & -\boldsymbol{\Sigma}_e[1,0] \\
    -\boldsymbol{\Sigma}_e[1,0] & \boldsymbol{\Sigma}_e[0,0]
    \end{bmatrix}
\end{equation}
where 
\begin{equation}
    \boldsymbol{\Sigma}_{2d} = 
    \begin{bmatrix}
    \boldsymbol{\Sigma}_e[0,0] & \boldsymbol{\Sigma}_e[1,0] \\
    \boldsymbol{\Sigma}_e[1,0] & \boldsymbol{\Sigma}_e[1,1]
    \end{bmatrix}.
\end{equation}

\noindent \textbf{Rendered intensity:}
\begin{equation}
    \tilde{\alpha} = \alpha \cdot \text{opacity}_r \cdot \frac{1}{\sqrt{\det (\boldsymbol{\Sigma}_{2d})}}.
\end{equation}

These parameters form the basis for backward propagation, linking 2D rendering to 3D Gaussian primitives.

\subsection{Backward Propagation}

\noindent Gradient from 2D mean:
\begin{equation}
    \frac{dL}{d \boldsymbol{\mu}_e} = \begin{bmatrix}
    \frac{dL}{d \boldsymbol{\mu}_{2d}[0]} \\
    \frac{dL}{d \boldsymbol{\mu}_{2d}[1]} \\
    0
    \end{bmatrix}.
\end{equation}

\noindent Gradients from rendered intensity:
\begin{equation}
    \frac{dL}{d \alpha} = \frac{dL}{d \tilde{\alpha}} \cdot \frac{\text{opacity}_{r}}{\sqrt{\det(\boldsymbol{\Sigma}_{2d})}},
\end{equation}
\begin{equation}
    \frac{d L}{d \text{opacity}_{r}} = \frac{d L}{d \tilde{\alpha}} \cdot \frac{\alpha}{\sqrt{\det(\boldsymbol{\Sigma}_{2d})}},
\end{equation}
\begin{equation}
    \frac{d L}{d \det(\boldsymbol{\Sigma}_{2d})} = \frac{d L}{d \tilde{\alpha}} \cdot \left( -\frac{1}{2} \cdot \frac{\alpha \times \text{opacity}_{r}}{[\det(\boldsymbol{\Sigma}_{2d})]^{3/2}} \right).
\end{equation}

These compute contributions from rendered intensity, essential for focus-aware rendering.

\noindent Gradient from $\boldsymbol{\Sigma}_{2d}^{-1}$ to $\boldsymbol{\Sigma}_{2d}$ (using $\boldsymbol{\Sigma}_{2d} \cdot \boldsymbol{\Sigma}_{2d}^{-1} = \boldsymbol{I}$):
\begin{equation}
    \frac{d L}{d \boldsymbol{\Sigma}_{2d}} = -\boldsymbol{\Sigma}_{2d} \cdot \frac{d L}{d \boldsymbol{\Sigma}_{2d}^{-1}} \cdot \boldsymbol{\Sigma}_{2d}
    + \frac{d L}{d \det(\boldsymbol{\Sigma}_{2d})} \cdot \det(\boldsymbol{\Sigma}_{2d}) \cdot \boldsymbol{\Sigma}_{2d}^{-1}.
\end{equation}

This combines direct inverse propagation with determinant contribution, ensuring covariance consistency.

\noindent Map to $\boldsymbol{\Sigma}_e$:
\begin{equation}
    \frac{d L}{d \boldsymbol{\Sigma}_e} = \begin{bmatrix}
    \frac{d L}{d \boldsymbol{\Sigma}_{2d}[0,0]} & \frac{d L}{d \boldsymbol{\Sigma}_{2d}[1,0]} & 0 \\
    \frac{d L}{d \boldsymbol{\Sigma}_{2d}[1,0]} & \frac{d L}{d \boldsymbol{\Sigma}_{2d}[1,1]} & 0 \\
    0 & 0 & 0
    \end{bmatrix}.
\end{equation}

\noindent From $\boldsymbol{\mu}_e$ and $\boldsymbol{\Sigma}_e$ to $\boldsymbol{\mu}_c$ and $\boldsymbol{\Sigma}_c$ (with $q = \boldsymbol{\mu}_c^{\top} \boldsymbol{\Sigma}_c^{-1} \boldsymbol{\mu}_c - \boldsymbol{\mu}_e^{\top} \boldsymbol{\Sigma}_e^{-1} \boldsymbol{\mu}_e$):
\begin{equation}
    \frac{d L}{d q} = \frac{d L}{d \text{opacity}_{r}} \cdot \left( -\frac{1}{2} \text{opacity}_{r} \right),
\end{equation}
\begin{equation}
    \frac{d L}{d \boldsymbol{\mu}_c} = \left( \boldsymbol{\Sigma}_e \cdot \boldsymbol{\Sigma}_c^{-1} \right)^{\top} \cdot \frac{d L}{d \boldsymbol{\mu}_e}
    + \frac{d L}{d q} \cdot 2 \boldsymbol{\Sigma}_c^{-1} \boldsymbol{\mu}_c,
\end{equation}
\begin{equation}
    \frac{d L}{d \boldsymbol{\Sigma}_c^{-1}} = \frac{d L}{d q} \cdot \boldsymbol{\mu}_c \boldsymbol{\mu}_c^{\top}
    + \frac{d L}{d \boldsymbol{\Sigma}_e^{-1}},
\end{equation}
\begin{equation}
    \frac{d L}{d \boldsymbol{\Sigma}_e^{-1}} = \frac{d L}{d q} \cdot (-\boldsymbol{\mu}_e \boldsymbol{\mu}_e^{\top})
    - \boldsymbol{\Sigma}_e \cdot \frac{d L}{d \boldsymbol{\Sigma}_e} \cdot \boldsymbol{\Sigma}_e.
\end{equation}

\noindent From $\boldsymbol{\Sigma}_c^{-1}$ to $\boldsymbol{\Sigma}_c$:
\begin{equation}
    \frac{d L}{d \boldsymbol{\Sigma}_c} = -\boldsymbol{\Sigma}_c^{-1} \cdot \frac{d L}{d \boldsymbol{\Sigma}_c^{-1}} \cdot \boldsymbol{\Sigma}_c^{-1}.
\end{equation}

This propagates gradients back to camera coordinates, maintaining differentiability.

\noindent Finally back to world coordinates:
\begin{equation}
    \frac{d L}{d \boldsymbol{\tilde{\mu}}} = R_c^{\top} \cdot \frac{d L}{d \boldsymbol{\mu}_c},
\end{equation}
\begin{equation}
    \frac{d L}{d \boldsymbol{\tilde{\Sigma}}} = R_c^{\top} \cdot \frac{d L}{d \boldsymbol{\Sigma}_c} \cdot R_c.
\end{equation}

Above final steps transform gradients to world space, enabling end-to-end optimization of 3D Gaussians.

\section{Implementation details of baseline methods}
Comparisons are conducted with various compression and 3D reconstruction methods. The traditional compression algorithm HEVC~\cite{sze2014high} is employed with a CRF value of $16$. For the SOTA NeRF-based method represented by INIF~\cite{dai2025implicit}, a SIREN-like neural network with frequency-adjusted activations is trained for $40000$ iterations using the Adam optimizer and a compression ratio of $16$, with automatic batch size tuning and JAX for GPU acceleration. The implementation of INIF uses its official code with default hyperparameters. Inspired by FruitNinja~\cite{wu2025fruitninja}, we also develop a slice-adapted version of original 3DGS~\cite{kerbl20233d} for evaluation. Two modifications are made: (i) \textbf{Internal initialization}: we employ grid-based ray-casting to fill interior voxels with Gaussian primitives, ensuring volumetric coverage beyond surface points; (ii) \textbf{Slice rendering}: unlike FruitNinja's infinitesimally thin cutting planes, we render slices with finite thickness by first masking out Gaussians whose distance from the imaging plane exceeds their spatial extent, then projecting the retained primitives to 2D Gaussian ellipses with depth-dependent attenuation. Contributions are additively accumulated to match intensity integration in slice imaging. All methods are run on NVIDIA A800 GPUs.

\section{More qualitative results}

\begin{figure*}[!ht]
\centerline{\includegraphics[width=17.4cm]{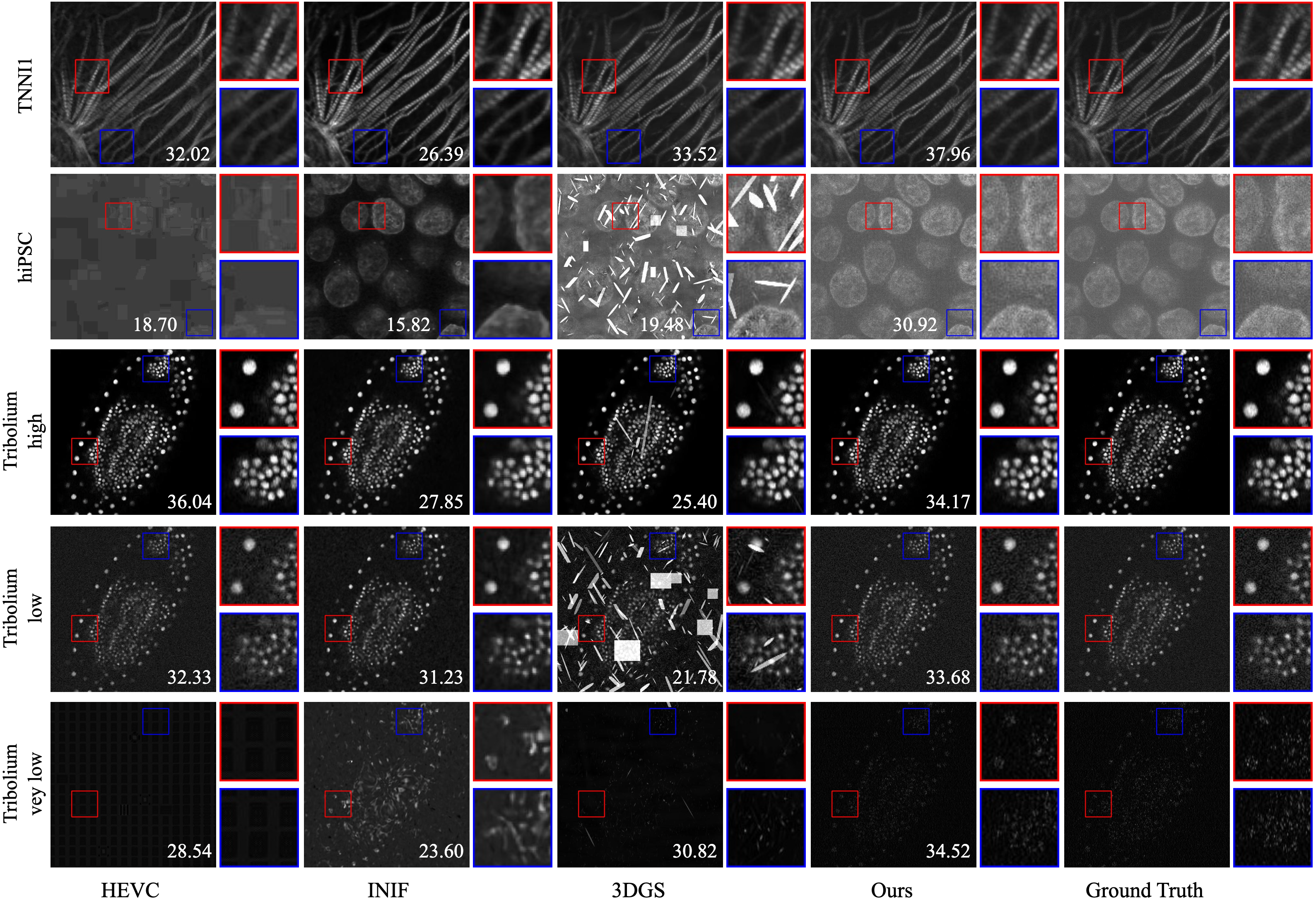}}
\caption{Qualitative comparison of reconstruction results on three cellular datasets. Slice examples of different methods with PSNR (dB) shown at the bottom of each image.}
\label{FIG_suppl_2d_quali_comparison}
\end{figure*}

\noindent \textbf{Main results.} We visualize more reconstruction results in Fig.~\ref{FIG_suppl_2d_quali_comparison}. HEVC and INIF introduce strong blocky or oversmoothed artifacts, failing to recover fine cellular structures faithfully. Original 3DGS preserves some details in high signal-to-noise conditions, but produces floating artifacts in low-signal conditions. In contrast, our method successfully recovers sharper morphology with cleaner backgrounds and better local contrast across all conditions, from high-quality TNNI1 to very-low-signal Tribolium dataset, closely matching the ground truth.

\begin{figure*}[!t]
\centerline{\includegraphics[width=16.4cm]{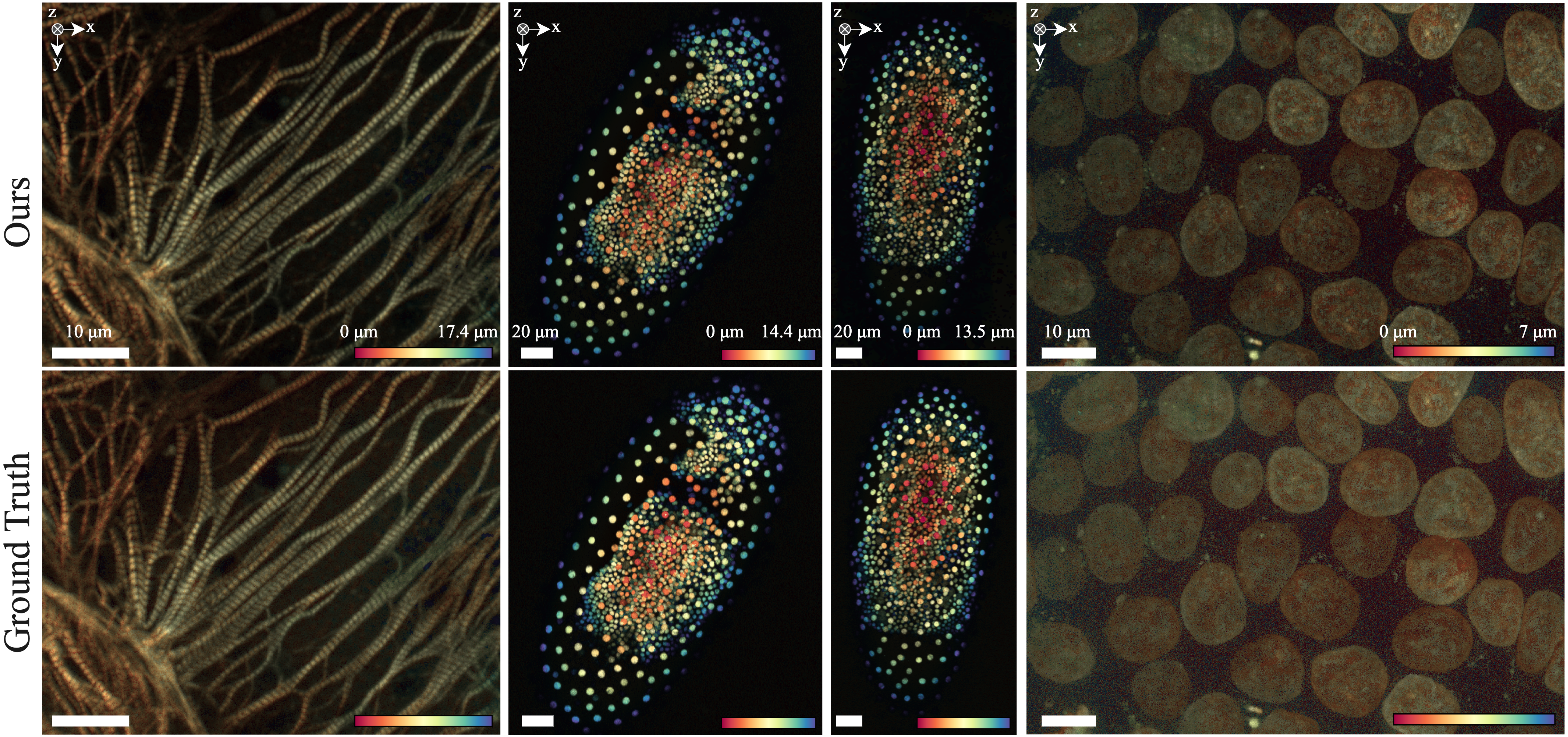}}
\caption{3D reconstruction results on cellular datasets, presented by depth-encoded color rendering of volumes.}
\label{FIG_suppl_3d_quali}
\end{figure*}

\noindent \textbf{Volume visualization.} In the main text, the 3D reconstruction results are presented as maximum-intensity projections (MIP), which appear in grayscale. To provide a clearer visualization of the volumetric geometry, we additionally show depth-colored renderings of the reconstructed volumes in Fig.~\ref{FIG_suppl_3d_quali}. The depth of the dominant signal along the z-axis is encoded using a perceptually uniform colormap, allowing fine-scale structural variations to be more easily perceived. This visualization is intended to improve perceptual understanding of the reconstructed 3D geometry.

\begin{figure}[!t]
\centerline{\includegraphics[width=\columnwidth]{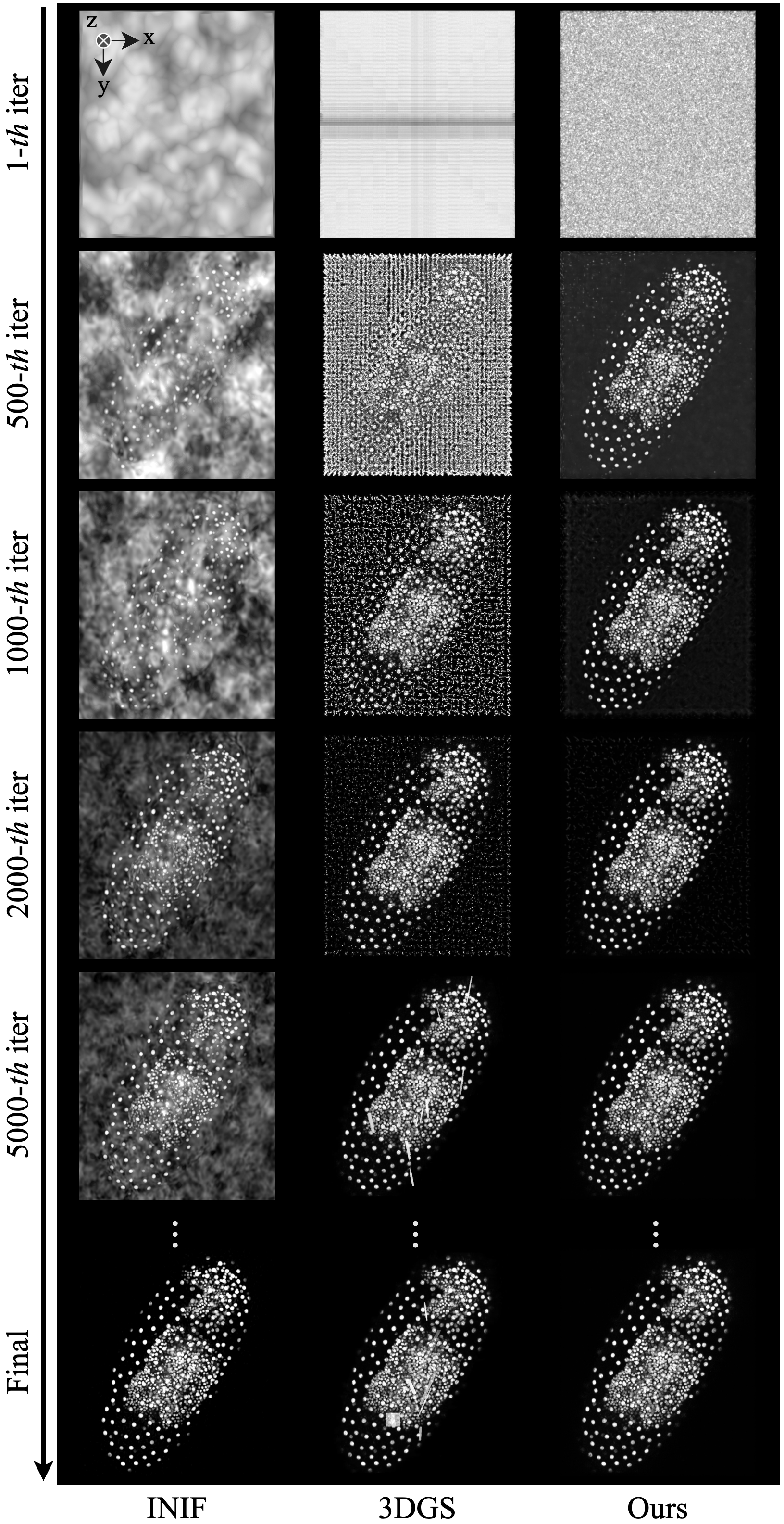}}
\caption{Convergence analysis of baseline methods and GaussianPile. We visualize the intermediate snapshots at different iterations. GaussianPile shows faster and better convergence.}
\label{FIG_suppl_convergence}
\vspace{-13pt}
\end{figure}

\noindent \textbf{Convergence analysis.} We conduct a visual convergence analysis of the INIF~\cite{dai2025implicit}, original 3DGS~\cite{kerbl20233d}, and GaussianPile in Fig.~\ref{FIG_suppl_convergence}. INIF gradually forms the structure but converges slowly. Original 3DGS converges faster than INIF, but struggles to cull the dense grid-pattern Gaussians in early iterations and subsequently exhibits persistent floating artifacts. In contrast, GaussianPile rapidly forms a compact and well-localized representation within $\sim2$k iterations. These observations demonstrate that GaussianPile achieves faster and more stable convergence while producing cleaner reconstructions.

\begin{figure}[!t]
\centerline{\includegraphics[width=\columnwidth]{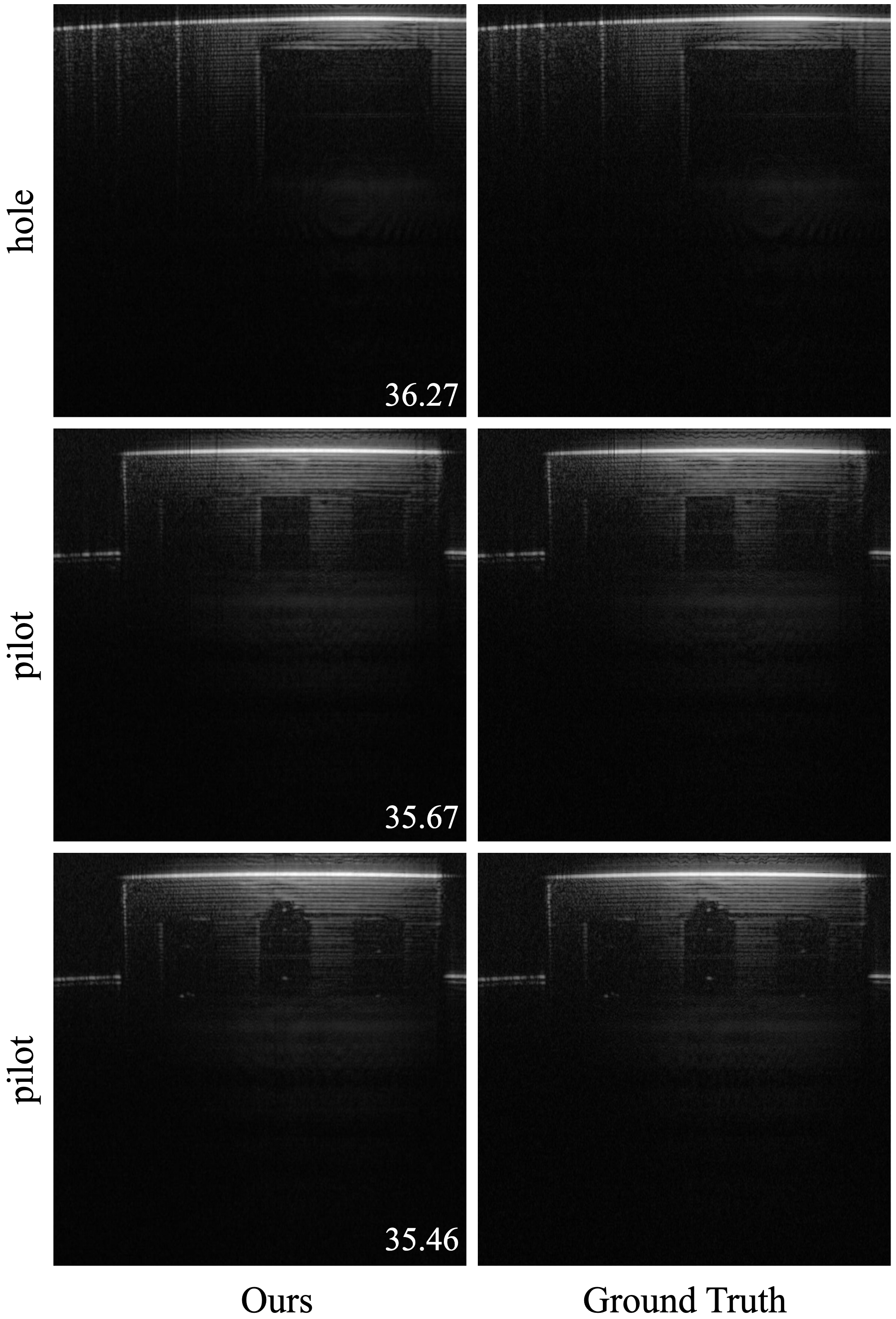}}
\caption{Qualitative results on industrial CeraMIRScan MIR-OCT data, validating GaussianPile's applicability to diverse slice-based imaging modalities.}
\label{FIG_suppl_quali_NDT}
\vspace{-5pt}
\end{figure}

\noindent \textbf{Application on Industrial Non-Destructive Testing.} To further assess the generality of GaussianPile, we evaluate it on the CeraMIRScan Mid-infrared Optical Coherence Tomography (MIR-OCT) dataset, a challenging industrial non-destructive testing (NDT) benchmark comprising $29$ MIR-OCT volumes of 3D-printed ceramic components. Unlike biomedical ultrasound and microscopy data used in our main text, MIR-OCT exhibits distinct noise characteristics, signal attenuation patterns, and structural textures, making it an ideal testbed for robustness. As shown in Fig.~\ref{FIG_suppl_quali_NDT}, GaussianPile successfully reconstructs the layered ceramic structures and internal patterns across samples (“hole”, “pilot”), producing results that closely match the ground-truth MIR-OCT slices. These results demonstrate that GaussianPile is not limited to biomedical 3D tomography and can be effectively applied to industrial inspection scenarios, highlighting its versatility for broader NDT applications.

\begin{figure*}[!t]
\centerline{\includegraphics[width=16.8cm]{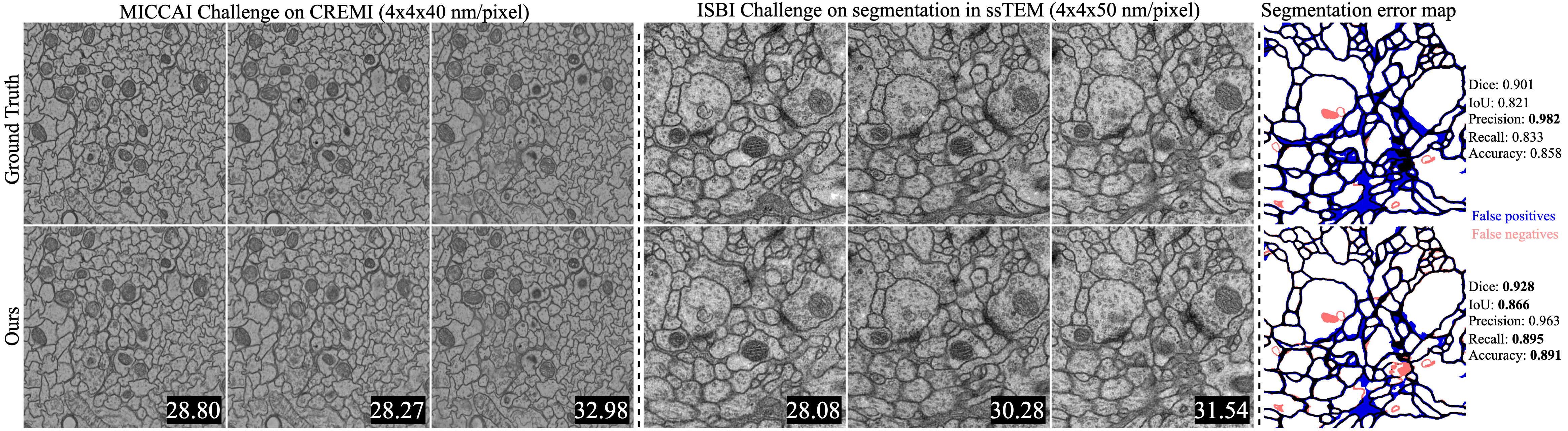}}
\caption{Generalization to highly anisotropic ssEM (CREMI\footref{fn:cremi}, ISBI12 \cite{arganda2015crowdsourcing}, $10\times/12.5\times$ anisotropy respectively). We visualize three consecutive slices for each dataset.}
\label{FIG_suppl_generalization}
\end{figure*}

\begin{figure*}[!h]
\centerline{\includegraphics[width=16.8cm]{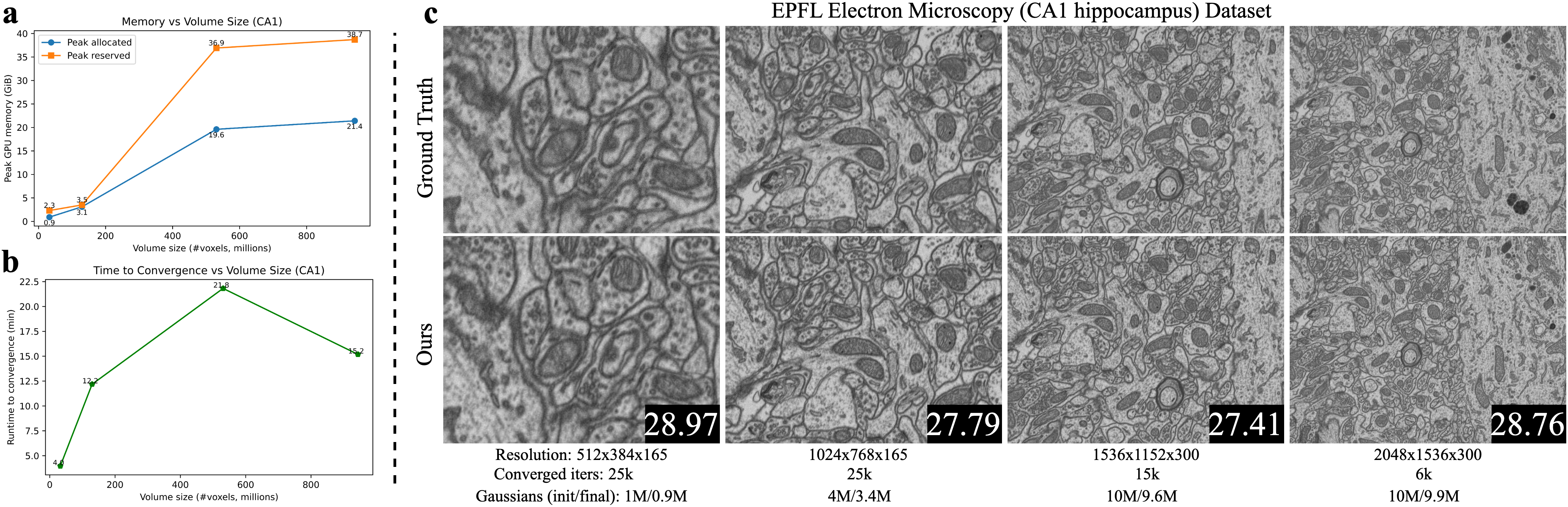}}
\caption{Scalability analysis on the large-scale EPFL CA1 EM dataset \cite{lucchi2011supervoxel}.}
\label{FIG_suppl_scalability}
\end{figure*}

\begin{table*}[!t]
\vspace{-4pt}
\centering
\scriptsize
\setlength{\tabcolsep}{3.2pt}
\renewcommand{\arraystretch}{0.95}
\caption{Quantitative comparison on highly anisotropic ssEM datasets. We report 2D slice fidelity, 3D volumetric fidelity, and compression/runtime metrics on ISBI12 \cite{arganda2015crowdsourcing} and CREMI\footref{fn:cremi}.}
\label{tab:suppl_ssem_quantitative}
\vspace{-2pt}
\begin{tabular}{@{}l c c c c c c c c c c c c@{}}
\toprule
\multirow{2}{*}{Method}
& \multicolumn{6}{c}{ISBI12}
& \multicolumn{6}{c}{CREMI} \\
\cmidrule(lr){2-7} \cmidrule(lr){8-13}
& 2D PSNR$\uparrow$ & 2D SSIM$\uparrow$ & 3D PSNR$\uparrow$ & 3D SSIM$\uparrow$ & CR$\uparrow$ & Time$\downarrow$
& 2D PSNR$\uparrow$ & 2D SSIM$\uparrow$ & 3D PSNR$\uparrow$ & 3D SSIM$\uparrow$ & CR$\uparrow$ & Time$\downarrow$ \\
\midrule
CoordNet~\cite{han2022coordnet}
& 25.19 & 0.708 & 24.98 & 0.715 & 3$\times$ & 51m
& 22.66 & 0.620 & 20.80 & 0.589 & 6$\times$ & 2h34m \\
NeurComp~\cite{lu2021compressive}
& 24.45 & 0.695 & 25.19 & 0.728 & \textbf{8}$\times$ & \textbf{5m}
& 19.81 & 0.428 & 17.62 & 0.333 & \textbf{10}$\times$ & 11h \\
Ours
& \textbf{28.79} & \textbf{0.815} & \textbf{28.98} & \textbf{0.881} & \textbf{8}$\times$ & 6m
& \textbf{29.50} & \textbf{0.831} & \textbf{28.27} & \textbf{0.752} & \textbf{10}$\times$ & \textbf{25m} \\
\bottomrule
\end{tabular}
\vspace{-8pt}
\end{table*}

\noindent \textbf{Generalization to highly anisotropic microscopy and task-driven evaluation.} We further evaluate GaussianPile on highly anisotropic serial-section electron microscopy (ssEM) datasets, including CREMI\footnote{\url{https://cremi.org/}\label{fn:cremi}} and ISBI12 \cite{arganda2015crowdsourcing}, where the axial resolution is much coarser than the in-plane resolution and the imaging process departs substantially from the Gaussian, spatially invariant PSF assumption. As shown in Fig.~\ref{FIG_suppl_generalization}, GaussianPile still preserves coherent slice structures and fine boundaries across consecutive sections, indicating robust generalization under substantial PSF mismatch and severe anisotropy. To complement PSNR/SSIM-style reconstruction metrics, we additionally report segmentation results on ISBI12 using a fixed 2D U-Net in the rightmost column. Segmentations obtained from our reconstructions exhibit improved Dice/IoU and fewer boundary errors, suggesting that GaussianPile better preserves task-relevant structural fidelity beyond voxel-level similarity.

\noindent \textbf{Scalability on large-scale volumes.} To further examine scalability, we evaluate GaussianPile on the large-scale EPFL CA1 EM volume, with resolutions up to $0.94\text{B}$ voxels. As shown in Fig.~\ref{FIG_suppl_scalability}, GaussianPile scales to near-billion-voxel data on a single A800 40GB GPU while maintaining stable reconstruction quality. Notably, the memory and runtime costs are governed primarily by the number of optimized Gaussian primitives rather than growing linearly with voxel resolution, highlighting the practicality of the representation for large volumes.

\section{More quantitative results}
To complement the qualitative generalization results on highly anisotropic ssEM data, we further report quantitative comparisons on ISBI12 \cite{arganda2015crowdsourcing} and CREMI\footref{fn:cremi} against representative INR baselines, including CoordNet~\cite{han2022coordnet} and NeurComp~\cite{lu2021compressive}. We summarize 2D slice fidelity (PSNR/SSIM), 3D volumetric fidelity (PSNR/SSIM), and compression efficiency/runtime (CR/Time) in Tab.~\ref{tab:suppl_ssem_quantitative}. These results further confirm that GaussianPile consistently outperforms the INR baselines on both reconstruction quality and practical efficiency for challenging anisotropic volumes. In particular, the gains on ISBI12 and CREMI support the qualitative observations and the segmentation results shown in Fig.~\ref{FIG_suppl_generalization}.

\end{document}


\title{GaussianPile: A Unified Sparse Gaussian Splatting Framework for Slice-based Volumetric Reconstruction \\ Supplementary Material}  

\maketitle
\thispagestyle{empty}
\appendix

\section{Derivation of Focus Gaussian Function}

Suppose $ g(\mathbf{x}) $ is a 3-dimensional Gaussian distribution with mean $\mathbf{\mu}$ and covariance matrix $\mathbf{\Sigma}$:
\begin{equation}
    g(\mathbf{x}) = \frac{a}{(2\pi)^{3/2} \sqrt{\det(\mathbf{\Sigma})}} \exp\left(-\frac{1}{2} (\mathbf{x} - \mathbf{\mu})^T \mathbf{\Sigma}^{-1} (\mathbf{x} - \mathbf{\mu})\right)
\end{equation}

\noindent where $ a $ is a normalization constant (typically $ a = 1 $ for a probability density), $\mathbf{x} = [x_1, x_2, x_3]^T$, and $\mathbf{\Sigma}$ is a 3×3 positive definite covariance matrix.

The weighting function $ h(\mathbf{x}) $ depends only on the third coordinate $ z = x_3 $, and we define it as:
\begin{equation}
    h(\mathbf{x}) = \exp\left(-\frac{z^2}{2 \sigma_z^2}\right)
\end{equation}

\noindent where $z = \mathbf{e}_3^T \mathbf{x} $, $\mathbf{e}_3 = [0, 0, 1]^T$, and $\sigma_z^2$ is a variance parameter. This $ h(\mathbf{x}) $ resembles the exponential part of a Gaussian in the $z$-direction.

The target function is:
\begin{equation}
    f(\mathbf{x}) = g(\mathbf{x}) \cdot h(\mathbf{x})
\end{equation}

Since $ f(\mathbf{x}) $ is the product of two terms with exponential forms, and we expect $ f(\mathbf{x}) $ to be Gaussian (as implied by the query specifying $\mathbf{\Sigma}_f^{-1}$ and $\mathbf{\mu}_f$), we aim to express $ f(\mathbf{x}) $ as:
\begin{equation}
    f(\mathbf{x}) \propto \exp\left(-\frac{1}{2} (\mathbf{x} - \mathbf{\mu}_f)^T \mathbf{\Sigma}_f^{-1} (\mathbf{x} - \mathbf{\mu}_f)\right)
\end{equation}

First, compute the exponent of $f(\mathbf{x}) = g(\mathbf{x}) \cdot h(\mathbf{x}) $. The exponent of $g(\mathbf{x})$ is:
\begin{equation}
    -\frac{1}{2} (\mathbf{x} - \mathbf{\mu})^T \mathbf{\Sigma}^{-1} (\mathbf{x} - \mathbf{\mu})
\end{equation}

The exponent of $h(\mathbf{x})$ is:
\begin{equation}
    -\frac{z^2}{2 \sigma_z^2}
\end{equation}

Since $z = \mathbf{e}_3^T \mathbf{x}$, we rewrite:
\begin{equation}
    z^2 = (\mathbf{e}_3^T \mathbf{x})^2 = (\mathbf{e}_3^T \mathbf{x})(\mathbf{x}^T \mathbf{e}_3) = \mathbf{x}^T (\mathbf{e}_3 \mathbf{e}_3^T) \mathbf{x}
\end{equation}

Thus:
\begin{equation}
    -\frac{z^2}{2 \sigma_z^2} = -\frac{1}{2 \sigma_z^2} \mathbf{x}^T (\mathbf{e}_3 \mathbf{e}_3^T) \mathbf{x} = -\frac{1}{2} \mathbf{x}^T \left( \frac{\mathbf{e}_3 \mathbf{e}_3^T}{\sigma_z^2} \right) \mathbf{x}
\end{equation}

The total exponent of $f(\mathbf{x})$ is:
\begin{equation}
    -\frac{1}{2} (\mathbf{x} - \mathbf{\mu})^T \mathbf{\Sigma}^{-1} (\mathbf{x} - \mathbf{\mu}) - \frac{1}{2} \mathbf{x}^T \left( \frac{\mathbf{e}_3 \mathbf{e}_3^T}{\sigma_z^2} \right) \mathbf{x}
\end{equation}

Expand the first term:
\begin{equation}
    (\mathbf{x} - \mathbf{\mu})^T \mathbf{\Sigma}^{-1} (\mathbf{x} - \mathbf{\mu}) = \mathbf{x}^T \mathbf{\Sigma}^{-1} \mathbf{x} - 2 \mathbf{x}^T \mathbf{\Sigma}^{-1} \mathbf{\mu} + \mathbf{\mu}^T \mathbf{\Sigma}^{-1} \mathbf{\mu}
\end{equation}

So the exponent becomes:
\begin{equation}
    -\frac{1}{2} \left[ \mathbf{x}^T \mathbf{\Sigma}^{-1} \mathbf{x} - 2 \mathbf{x}^T \mathbf{\Sigma}^{-1} \mathbf{\mu} + \mathbf{\mu}^T \mathbf{\Sigma}^{-1} \mathbf{\mu} + \mathbf{x}^T \left( \frac{\mathbf{e}_3 \mathbf{e}_3^T}{\sigma_z^2} \right) \mathbf{x} \right]
\end{equation}

Combine the quadratic terms:
\begin{equation}
    \mathbf{x}^T \mathbf{\Sigma}^{-1} \mathbf{x} + \mathbf{x}^T \left( \frac{\mathbf{e}_3 \mathbf{e}_3^T}{\sigma_z^2} \right) \mathbf{x} = \mathbf{x}^T \left( \mathbf{\Sigma}^{-1} + \frac{\mathbf{e}_3 \mathbf{e}_3^T}{\sigma_z^2} \right) \mathbf{x}
\end{equation}

The exponent is now:
\begin{equation}
    -\frac{1}{2} \left[ \mathbf{x}^T \left( \mathbf{\Sigma}^{-1} + \frac{\mathbf{e}_3 \mathbf{e}_3^T}{\sigma_z^2} \right) \mathbf{x} - 2 \mathbf{x}^T \mathbf{\Sigma}^{-1} \mathbf{\mu} + \mathbf{\mu}^T \mathbf{\Sigma}^{-1} \mathbf{\mu} \right]
\end{equation}

We want this to match the form $-\frac{1}{2} (\mathbf{x} - \mathbf{\mu}_f)^T \mathbf{\Sigma}_f^{-1} (\mathbf{x} - \mathbf{\mu}_f)$. Expand the target form:
\begin{equation}
    (\mathbf{x} - \mathbf{\mu}_f)^T \mathbf{\Sigma}_f^{-1} (\mathbf{x} - \mathbf{\mu}_f) = \mathbf{x}^T \mathbf{\Sigma}_f^{-1} \mathbf{x} - 2 \mathbf{x}^T \mathbf{\Sigma}_f^{-1} \mathbf{\mu}_f + \mathbf{\mu}_f^T \mathbf{\Sigma}_f^{-1} \mathbf{\mu}_f
\end{equation}

Compare coefficients with our exponent:\\
\noindent \textbf{Quadratic term:} $\mathbf{x}^T \left( \mathbf{\Sigma}^{-1} + \frac{\mathbf{e}_3 \mathbf{e}_3^T}{\sigma_z^2} \right) \mathbf{x} = \mathbf{x}^T \mathbf{\Sigma}_f^{-1} \mathbf{x}$
\begin{equation}
    \mathbf{\Sigma}_f^{-1} = \mathbf{\Sigma}^{-1} + \frac{\mathbf{e}_3 \mathbf{e}_3^T}{\sigma_z^2}
\end{equation}

Noting that $\frac{1}{\sigma_z \sigma_z} = \frac{1}{\sigma_z^2}$, and $\mathbf{e}_3 \mathbf{e}_3^T = [0,0,1]^T [0,0,1]$.

\noindent \textbf{Linear term:} $-2 \mathbf{x}^T \mathbf{\Sigma}^{-1} \mathbf{\mu} = -2 \mathbf{x}^T \mathbf{\Sigma}_f^{-1} \mathbf{\mu}_f$
\begin{equation}
    \mathbf{\Sigma}_f^{-1} \mathbf{\mu}_f = \mathbf{\Sigma}^{-1} \mathbf{\mu}
\end{equation}

Solve for $\mathbf{\mu}_f$:
\begin{equation}
    \mathbf{\mu}_f = \mathbf{\Sigma}_f \mathbf{\Sigma}^{-1} \mathbf{\mu}
\end{equation}

(Here, $\mathbf{\Sigma}_f = (\mathbf{\Sigma}_f^{-1})^{-1}$, which is consistent since $\mathbf{\Sigma}_f^{-1}$ is invertible.)

\noindent \textbf{Constant term:} The remaining term $\mathbf{\mu}^T \mathbf{\Sigma}^{-1} \mathbf{\mu} - \mathbf{\mu}_f^T \mathbf{\Sigma}_f^{-1} \mathbf{\mu}_f$ affects the normalization constant.

The exponent becomes:
\begin{equation}
    -\frac{1}{2} \left[ \mathbf{x}^T \mathbf{\Sigma}_f^{-1} \mathbf{x} - 2 \mathbf{x}^T \mathbf{\Sigma}_f^{-1} \mathbf{\mu}_f + \mathbf{\mu}^T \mathbf{\Sigma}^{-1} \mathbf{\mu} \right]
\end{equation}

Complete the square:
\begin{equation}
    \mathbf{x}^T \mathbf{\Sigma}_f^{-1} \mathbf{x} - 2 \mathbf{x}^T \mathbf{\Sigma}_f^{-1} \mathbf{\mu}_f = (\mathbf{x} - \mathbf{\mu}_f)^T \mathbf{\Sigma}_f^{-1} (\mathbf{x} - \mathbf{\mu}_f) - \mathbf{\mu}_f^T \mathbf{\Sigma}_f^{-1} \mathbf{\mu}_f
\end{equation}

This confirms that $ f(\mathbf{x}) $ is a Gaussian function with mean $\mathbf{\mu}_f$ and precision $\mathbf{\Sigma}_f^{-1}$.


\subsection{Response length}
Author responses must be no longer than 1 page in length including any references and figures.
Overlength responses will simply not be reviewed.
This includes responses where the margins and formatting are deemed to have been significantly altered from those laid down by this style guide.
Note that this \LaTeX\ guide already sets figure captions and references in a smaller font.

\section{Formatting your Response}

{\bf Make sure to update the paper title and paper ID in the appropriate place in the tex file.}

All text must be in a two-column format.
The total allowable size of the text area is $6\frac78$ inches (17.46 cm) wide by $8\frac78$ inches (22.54 cm) high.
Columns are to be $3\frac14$ inches (8.25 cm) wide, with a $\frac{5}{16}$ inch (0.8 cm) space between them.
The top margin should begin 1 inch (2.54 cm) from the top edge of the page.
The bottom margin should be $1\frac{1}{8}$ inches (2.86 cm) from the bottom edge of the page for $8.5 \times 11$-inch paper;
for A4 paper, approximately $1\frac{5}{8}$ inches (4.13 cm) from the bottom edge of the page.

Please number any displayed equations.
It is important for readers to be able to refer to any particular equation.

Wherever Times is specified, Times Roman may also be used.
Main text should be in 10-point Times, single-spaced.
Section headings should be in 10 or 12 point Times.
All paragraphs should be indented 1 pica (approx.~$\frac{1}{6}$ inch or 0.422 cm).
Figure and table captions should be 9-point Roman type as in \cref{fig:onecol}.

List and number all bibliographical references in 9-point Times, single-spaced,
at the end of your response.
When referenced in the text, enclose the citation number in square brackets, for example~\cite{Alpher05}.
Where appropriate, include the name(s) of editors of referenced books.

\begin{figure}[t]
  \centering
  \fbox{\rule{0pt}{0.5in} \rule{0.9\linewidth}{0pt}}
   \caption{Example of caption.  It is set in Roman so that mathematics
   (always set in Roman: $B \sin A = A \sin B$) may be included without an
   ugly clash.}
   \label{fig:onecol}
\end{figure}

To avoid ambiguities, it is best if the numbering for equations, figures, tables, and references in the author response does not overlap with that in the main paper (the reviewer may wonder if you talk about \cref{fig:onecol} in the author response or in the paper).
See \LaTeX\ template for a workaround.

\subsection{Illustrations, graphs, and photographs}

All graphics should be centered.
Please ensure that any point you wish to make is resolvable in a printed copy of the response.
Resize fonts in figures to match the font in the body text, and choose line widths which render effectively in print.
Readers (and reviewers), even of an electronic copy, may choose to print your response in order to read it.
You cannot insist that they do otherwise, and therefore must not assume that they can zoom in to see tiny details on a graphic.

When placing figures in \LaTeX, it is almost always best to use \verb+\includegraphics+, and to specify the  figure width as a multiple of the line width as in the example below
{\small\begin{verbatim}
   \usepackage{graphicx} ...
   \includegraphics[width=0.8\linewidth]
                   {myfile.pdf}
\end{verbatim}
}

{
    \small
    \bibliographystyle{ieeenat_fullname}
    \bibliography{main}
}